%% file: main.tex
\documentclass[acmsmall,nonacm]{acmart}

\AtBeginDocument{%
  \providecommand\BibTeX{{%
    \normalfont B\kern-0.5em{\scshape i\kern-0.25em b}\kern-0.8em\TeX}}}

\usepackage{amsmath,amsfonts}
\usepackage{array}
\usepackage{textcomp}
\usepackage{stfloats}
\usepackage{url}
\usepackage{verbatim}
\usepackage{graphicx}
\usepackage{import}
\usepackage{color,soul}
\usepackage{float}
\usepackage[export]{adjustbox}
\usepackage{subcaption}
\usepackage{tikz}
\definecolor[named]{Maroon}{cmyk}{0,1,1,0.5} % Can't import xcolor again - defining maroon here
\DeclareRobustCommand*\circled[1]{\tikz[baseline=(char.base)]{
        \node[shape=circle,draw,inner sep=1.5pt, Maroon, fill=Maroon] (char)
            {\color{white}\scriptsize\textbf{#1}};}
        }

\usepackage[linesnumbered,ruled,vlined]{algorithm2e}
\begin{document}

\title{HALO: Fault-Tolerant Safety Architecture For High-Speed Autonomous Racing}

\author{Aron Harder, Amar Kulkarni, and Madhur Behl \\ Department of Computer Science, University of Virginia \\ (ah2ph,madhur.behl,ark8su)@virginia.edu}

\renewcommand{\shortauthors}{Harder,A. et al.}

\begin{abstract}
\import{sections/}{abstract.tex}
\end{abstract}

\maketitle

\section{Introduction}
\import{sections/}{introduction.tex}

\section{Related Work}
\import{sections/}{related_work.tex}

\section{Autonomous Racing Setup}
\label{sec:setup}
\import{sections/}{setup.tex}
\section{Software Stack Overview}
\label{sec:overview}
\import{sections/}{overview.tex}

\section{Control Module}
\label{sec:control}
\import{sections/}{control_module.tex}

\section{Localization Module}
\label{sec:local}
\import{sections/}{localization_module.tex}

\section{Communication Module}
\label{sec:comm}
\import{sections/}{communication_module.tex}

\section{Perception Module}
\label{sec:perception}
\import{sections/}{perception_module.tex}

\section{Safety Analysis and Failure Modes}
\label{sec:safety}
\import{sections/}{safety_module.tex}

\section{HALO Safety Architecture}
\label{sec:halo}
\import{sections/}{safety_architecture.tex}

\section{Results and Evaluation}
\import{sections/}{results.tex}

\section{Discussion}
\import{sections/}{discussion.tex}

\section{Conclusion and Future Work}
\import{sections/}{conclusion.tex}

\bibliography{references.bib}{}
\bibliographystyle{IEEEtran}

\end{document}

%% file: sections/abstract.tex
The field of high-speed autonomous racing has seen significant advances in recent years, with the rise of competitions such as RoboRace and the Indy Autonomous Challenge providing a platform for researchers to develop software stacks for autonomous race vehicles capable of reaching speeds in excess of 170 mph. Ensuring the safety of these vehicles requires the software to continuously monitor for different faults and erroneous operating conditions during high-speed operation, with the goal of mitigating any unreasonable risks posed by malfunctions in sub-systems and components.
%While open-source self-driving stacks such as Autoware and Apollo exist, there has been a lack of comprehensive examination of the safety aspects of autonomous driving at both the programming and functional level. 
This paper presents a comprehensive overview of the HALO safety architecture, which has been implemented on a full-scale autonomous racing vehicle as part of the Indy Autonomous Challenge. The paper begins with a failure mode and criticality analysis of the perception, planning, control, and communication modules of the software stack. Specifically, we examine three different types of faults - node health, data health, and behavioral-safety faults. To mitigate these faults, the paper then outlines HALO safety archetypes and runtime monitoring methods. Finally, the paper demonstrates the effectiveness of the HALO safety architecture for each of the faults, through real-world data gathered from autonomous racing vehicle trials during multi-agent scenarios.

%% file: sections/introduction.tex
Motorsport racing is where new concepts and technologies can be proven and stressed to the breaking point long before production lines are established. Several of the safety features (disc brakes, anti-lock brakes,  rear view mirrors, LED headlights etc.) in our automobiles, that we take for granted, owe their origins to motorsport racing. 
It is worth remembering there was a time when we were transitioning from \textit{horse-driven} carriages to \textit{horse-less} carriages and people worried about cars themselves, and did not have trust in the merits, and safety of the technology. 
It was through racing events that car manufactures were able to showcase the safety of their vehicles.
We are now experiencing a similar transition, from \textit{driver-driven} vehicles to \textit{driver-less} vehicles and public concerns surrounding the safety of autonomous driving are ever increasing~\cite{aaa_survey}.

Autonomous racing, is therefore an ideal testing ground for the development of autonomous driving algorithms capable of mastering the most challenging and rare situations. The high-speed, close proximity nature of motorsport racing makes it suitable for learning agile autonomous driving behavior which can eventually inform motion planning for commercial autonomous self-driving and autonomous cyber-physical systems. 
Autonomous racing also presents a unique challenge to develop robust software that is capable of maintaining high performance and ensuring safety during aggressive maneuvers and speeds exceeding 150mph!

In October 2021, a historic autonomous race was held at the Indianapolis Motor Speedway (home to the Indy 500 race) as part of the Indy Autonomous Challenge (IAC)~\cite{iac}. 
Following in the footsteps of the DARPA Grand~\cite{buehler20072005} and Urban Challenges~\cite{buehler2009darpa} in 2004, and 2007, the IAC invited university teams to participate in a competitive autonomous race and showcase their research and ideas to the general public. 
This was full-scale, fully-autonomous open-wheel racing, a type of automobile racing characterized by the use of open wheels and open cockpits, which provides a unique set of aerodynamic, handling, and safety demands for autonomous racing systems.
Teams were asked to develop the complete autonomous racing software stack for the standardized racing vehicle Dallara AV-21, a modified Indy Lights vehicle, and race each other using a ruleset comparable to human drivers. Nine university teams from all over the world participated in this race. A subsequent race was held at the Las Vegas Motor Speedway track in January 2022. This time it was a head-to-head autonomous race and teams had to overtake an opponent at racing speeds.
All nine race cars were built the same, with no hardware advantage from one car to another, and no hardware modifications were allowed. 
This was purely a \textit{``battle of algorithms''} and a team's perception, planning, and controlstack determined their competitiveness.

Even though a racing circuit provides a less cluttered, and an ideal environment for the autonomous vehicle (e.g. we only race in good weather conditions), software errors can rapidly lead to situations that cause crashes and severe physical damage to the race cars, resulting in cost-and-time intensive repairs. 
The high-speed and complex nature of autonomous racing creates unique demands on the software that controls the vehicles. 
These demands stem from the need to ensure safety while pushing the performance limits of the vehicles. 
Traditional safety systems for autonomous vehicles often fall short under these extreme conditions--it is not enough to design control models that detail the physical limits of the vehicle. For any algorithm that is implemented, certain safety checks are required to ensure that it is functioning as desired.
The consequences of software errors on a racing circuit can be severe, with crashes leading to costly repairs and downtime.
For building a fast fully-autonomous software stack, teams need the ability to detect software issues both at the design time, but more importantly at runtime as well. 
%The race car not only needs to complete the race, but the software needs to continuously monitor the autonomous operation and bring the car to a safe stop in case of any kind of software failure. 

To address these challenges, we introduce HALO - a fault-tolerant safety architecture designed specifically for autonomous racing vehicles. 
HALO's goal is to keep the vehicle in safe operation while maintaining its full performance capabilities, by continuously monitoring and mitigating software errors in the perception, planning, control, and communication modules of the autonomous racing software stack. 
To this end, we conduct a failure mode and criticality analysis to identify potential safety-critical faults and provide examples of safety archetypes that can be used to mitigate these faults at runtime. 
These safety archetypes are not specific to our implementation and can be applied to other autonomous vehicles.
The name HALO is inspired from Formula One racing, where a Halo is a protective barrier titanium ring around the driver that helps to prevent large objects and debris from entering the cockpit of a single-seat racing car, thereby keeping the driver safe at all time. In our case, the HALO architecture keeps our "autonomous driver" (and hence the racecar) safe at all times.

This paper has three main contributions:
\begin{enumerate}
\item With this paper, we provide the first holistic overview of the software safety architecture for high-speed autonomous racing implemented on a real, full-scale autonomous racecar. We provide a description of the components of an autonomous racing stack, and specifically discuss how fault-tolerant measures were used to ensure safety for those components. 
% Several of the safety features could be translated into regular autonomous driving without loss of generality.  This is the first such paper to present such a deep-dive into real-world fault-tolerance and failure modes for a complete autonomous racing stack.
\item We conduct a failure mode, effects, and criticality analysis (FMECA) for the autonomous racing software stack by reviewing software modules, their components, assemblies, and subsystems to identify potential failure modes, their causes and effects, and their criticality. 
\item Our HALO framework, identifies common failure modes and proposes safety node archetypes for runtime monitoring that can mitigate data health, node health, and behavioral-safety faults that could occur at a sensor level, algorithmic level, or at a system level on the vehicle. 
\end{enumerate}

We demonstrate the effectiveness of the HALO framework using ROS data collected from real-world race car runs during the Indy Autonomous Challenge.  We describe the failure modes in a comprehensive manner and provide runtime monitoring architectures for handling these failure modes. 

%% file: sections/related_work.tex
% Paragraph 1: Open-Source Driving Stacks
% Paragraph 2: Autonomous Vehicle Safety Architectures
% Paragraph 3: CPS Fault Detection, Identification, and Isolation
% Paragraph 4: ROS2 Fault Detection, Identification, and Isolation

The current landscape of safety for fully autonomous vehicles is vast and spans across a wide spectrum. Here we focus on work that closely aligns with fault tolerance and safety architectures at code or functional level.
%There exist several open-source self-driving stacks, such as Baidu Apollo Open Platform \cite{apollo_open}, openpilot \cite{openpilot}, and Autoware Auto \cite{autoware_auto}.
%While these provide the functionality for vehicles to drive autonomously, they do not focus on fault tolerance in great detail.
%Safety features provided by Autoware only consider monitor computing resources during runtime but do not check on the health of the code or data~\cite{autoware_sys_monitor}.
Many works have already looked at providing fault tolerant frameworks for similar systems.
These works describe techniques for fault detection in cyber-physical systems generally and self-driving vehicles specifically, and can be applied to autonomous racing.
%In this section, we will present related works in three areas: (1) existing autonomous vehicle safety architectures, (2) fault detection for general cyber-physical systems, and (3) fault detection for ROS2.
In this section, we will present related works in four areas: (1) fault detection for cyber-physical systems, (2) autonomous racing software architectures, (3) existing autonomous vehicle safety architectures, and fault detection for ROS2.

\noindent \textbf{1) Fault Detection in Cyber-Physical Systems:}
Faults in cyber-physical systems is a well-studied area, and many techniques for fault detection and isolation have been proposed.
%Several techniques used for fault detection and isolation are based on methods used more generally for cyber-physical systems.
A general approach to classifying types of failures across diverse safety-critical applications is described in~\cite{architectural_principles}.
%This work also provides general solutions to mitigating these failures both during testing before the system is deployed, and during runtime operation.
This work also discusses voting planes as a solution to mitigating sensor failures. For the HALO framework, voting planes are not a viable solution, due to hardware limitations -- for most systems, the race car does not contain enough sensors to reach a consensus, and the IAC rules forbid making hardware changes to the vehicle.
\cite{fault_tolerance_control} studies fault-tolerant control systems, distinguishing between passive or active systems.
In passive systems it is known in advance what faults are possible, while active systems must actively monitor their subsystems for faults and mitigate them during runtime.
For systems in which the mathematical model is known, a common approach is to use a Kalman Filter as in~\cite{fault_detection_kalman}.
By contrast, ~\cite{model_free_detection} proposes a model-free approach for fault detection in large-scale cyber-physical systems such as smart buildings.
%This method uses sensor clustering and a k-mediods algorithm to detect faults in the absence of having a mathematical model for the data.
The downside of this model-free approach is that it requires many sensors to perform the clustering algorithm.
\cite{online_error_detection} proposes a deep learning approach to fault detection in sensor data, using an LSTM model.
Our work avoids deep learning approaches to safety due to their black-box nature that cannot provide any guarantees of safety.
While the previous works mostly discuss detection and isolation of hardware faults, we also need to have safety checks on our software.
% \cite{case_for_shm} describes the process for mitigating software health faults as having the same approach as hardware faults, requiring checks on the data output for and system performance.
% These are approaches that we implemented into the HALO safety system.
\cite{case_for_shm} describes the process for mitigating software health faults in four stages - detecting problems, diagnosis and disambiguation, prediction, and mitigation.
The HALO safety system implements each these four stages: it detects problems through analysis of data output from the software; it diagnoses which software has exhibited a fault for further analysis by a human operator; it predicts time-to-failure through the fault thresholds encoded within the HALO system; and it mitigates threats by switching to a more reliable data source, by preventing improper behavior in the case of a behavioral fault, or by bringing the vehicle to a stop if necessary.
\cite{architectural_principles} discusses watchdog timers to detect software failure, a safety technique that we implemented in the form of a software heartbeat.
%Although many of the fault detection and mitigation techniques exist, there are some domain-specific features of the HALO safety architecture, such as needing to obey flags from race control that require unique safety solutions.
Although the HALO system uses fault detection and mitigation techniques already described by these works, the autonomous racing domain required certain adaptations, such as the inclusion of requirements in the race rules, and the trade-off between safety and the risk necessary to win a race.
%The approaches proposed by these works are all approaches for general cyber-physical systems that could be applied to self-driving vehicles.

\noindent \textbf{2) Autonomous Vehicle Safety Architectures:}
Self-driving vehicles are a prime example of a cyber-physical system with many possible points of failure.
There exist several open-source self-driving stacks, such as Baidu Apollo Open Platform \cite{apollo_open}, openpilot \cite{openpilot}, and Autoware Auto \cite{autoware_auto}.
While these provide the functionality for vehicles to drive autonomously, they do not focus on fault tolerance in great detail.
Safety features provided by Autoware only consider monitor computing resources during runtime but do not check on the health of the code or data~\cite{autoware_sys_monitor}.
To fill the gaps left by the open-source platforms, Guardauto \cite{guardauto} proposes a runtime safety system for self-driving stacks.
%The Guardauto system implements three safety features. First, it implements a system isolation to group logical components, so that a failure in one component does not cascade to other components.
%Second, it implements a local self-protection for each component by analyzing the data to detect an anomaly, and when an anomaly is detected it then resets the component to a previous state.
%Third, it implements a global self-protection for all components, to detect and isolate anomalous data being sent between components.
The Guardauto system implements safety features to isolate components for protection, detect anomalies, and reject anomalous data.
While the Guardauto system considers defenses to prevent code failures and data failures, it does not discuss what the vehicle should do in the event of a failure during operation.
Research into fault tolerance for automobiles has examined detecting faulty sensors~\cite{SVD_fault_detection,multisensor_fault_detection}.
%uses numerical analysis and recursive singular value decomposition to detect sensor faults.
%In this approach, an ensemble learning scheme is used.
%In this ensemble approach, the real-time measurements, test statistics from a pool of local models, and processed test statistics are all compared combined to form a single anomaly detector.
These works show that both numerical analysis and deep learning approaches can be used to compare real-time measurements against test statistics to detect anomalies in sensor outputs.
%In contrast,  uses a neural network approach for fault detection and identification.
%The sensor signals are processed using a single 1D CNN to find features within the data.
%After this processing, a deep network is used for classification, to identify not only if the data is anomalous, but what kind of anomaly is present, between a slow drift in values, a sudden spike in values, and a large increase in noise.
%Features are processed from the sensor signals using a CNN, and then used in a classifier network to detect anomalies.
The authors of ~\cite{hybrid_approaches} monitor the expected speed and trajectory of an autonomous vehicle for deviations to detect faults.
For this type of analysis, a single erroneous measurement could still be acceptable due to sensor uncertainty, a Kalman filter is used to predict data trends rather than monitoring on individual data points.
These papers describe techniques for detecting faults in self-driving vehicles, but do not close the loop to describe what the vehicle should do if a fault occurs.
Some faults can be handled through sensor redundancy and allow the vehicle to continue operating, but some require immediate action to ensure the safety of the passengers. 
\cite{Safe_Motion_Planning} uses a game theory approach for safe path following. This paper focuses on a single aspect of safety (path following), and does not include interactions with other vehicles.
None of these papers provides a holistic approach to vehicle safety, or consider having different responses depending on the severity of the fault.

\noindent \textbf{3) Autonomous Racing Software Architectures:}
The significance of safety escalates in autonomous racing, where minute errors in vehicle control can yield disastrous consequences. 
Several software architectures employed in the Indy Autonomous Challenge have been put forth.
The TUM team, for instance has published their software architecture used in the challenge~\cite{IAC_TUM}. This paper describes the assumptions made when designing each software module, but omits any discussion on safety checks or fault detection mechanisms for identifying any violations of these assumptions.
KAIST's paper~\cite{IAC_KAIST} focuses on a resilient navigation system for localization on the racetrack, but falls short of addressing other crucial software elements or safety checks that supervise code health or vehicular behavior. 
Papers from EuroRacing and MIT-Pitt-RW teams~\cite{IAC_EURO,IAC_MIT} primarily concentrate on the motion planning and control architecture employed during the challenge, with safety considerations limited to model constraints.

These works emphasize the development of models that prevent the vehicle from surpassing physical parameters and avoiding collisions (which is indeed an aspect of safety) but neglect scenarios involving faulty sensor data or improper code implementation (which is the focus of this paper). 
Outside of the IAC, RoboRace is also concerned with safety in autonomous racing software.
These papers ~\cite{Stahl_RoboRace1,Stahl_RoboRace2} both address safety checks on the trajectory planner of a RoboRace vehicle. However, this implementation assumes perfect data, and their only response to a failure is to bring the vehicle to a stop rather than introducing redundancy.
% ~\cite{IAC_TUM} includes a full stack architecture used by a winning vehicle, while ~\cite{IAC_EURO} focuses the motion planning and control aspects of autonomous racing.
% Neither of these papers discuss safety beyond acknowledging the necessity of an emergency stopping procedure.
% ~\cite{IAC_KAIST} discusses a resilient navigation system, but does not discuss monitoring code or vehicle behavior.
% None of these papers discuss safety monitoring of the entire software stack of the vehicle, merely assuming that their safety models are implemented and functioning correctly. Additionally, only KAIST discusses a safety response other than immediately bringing the vehicle to a stop.
% %None of these papers discuss safety monitoring of the entire software stack of the vehicle, or discuss different responses to different system faults.
None of these papers sufficiently cover the safety monitoring of the entire software stack of the vehicle, and there is an implicit assumption that their autonomous vehicles modules are correctly implemented and functioning. 
In this paper, we aim to address this gap by proposing comprehensive safety measures that go beyond functional safety, considering potential faults in sensor data and software implementation, thus contributing to safer and more reliable autonomous racing software.

\noindent \textbf{4) Fault Detection in ROS2:}
In addition to the faults discussed above, the Robot Operating System (ROS) used by open-source driving stacks has also been studied from a fault detection and isolation perspective.
% \cite{ROS_rescue} looks at fault tolerance of the ROS language itself. ROS uses a node architecture, with a single ROS master that acts as a registration and lookup service for nodes, parameters, and services.
ROS1 uses an architecture, with a single ROS master that acts as a registration and lookup service for nodes, parameters, and services.
\cite{ROS_rescue} provides fault tolerance of the ROS by providing a backup system for the central point of failure - i.e. the ROS master.
% The ROS master therefore presents a single point of failure for the language.
% \cite{ROS_rescue} proposes a system called ROS Rescue which provides a backup log of the ROS master.
% In the event that the ROS master has a critical error, the ROS Rescue system will reboot the ROS master, loading its data from the backup log.
% \cite{performance_ros2} explores the latency caused by converting data between the formats expected by the ROS2 nodes and those expected by the Data Distribution Service (DDS).
ROS2 provides a more robust architecture that does not rely on a single point of failure, but still has known issues with timing and messaging~\cite{performance_ros2}.
%explores the latency caused by converting data between the formats expected by the ROS2 nodes and those expected by the Data Distribution Service (DDS) which is responsible for distributing the data to all ROS2 nodes.
% This paper concludes that the latency caused by DDS is outweighed by the fault tolerance benefits it provides.
% The fault tolerance benefits of DDS include being able to save past time through a Quality of Service profile, and that DDS makes a ROS master unnecessary.
% \cite{robotic_anomaly_detection} explores using analysis of the timing of ROS messages in a system for anomaly detection. The system this paper implements looks at the rate of messages per topic, the rate of messages per time of day, and how often a message of one type tends to follow a message of another type.
% \cite{robotic_anomaly_detection} also proposes recommendations for improvements to the ROS2 language to make anomaly detection easier.
% These recommendations include strict value ranges for data publishing, a node state tracking and recovery system, and a safe execution mode for ROS nodes.
\cite{robotic_anomaly_detection} attempts to detect anomalies by analyzing the timing of ROS2 messages in a system - a sudden change in the publishing rate of a message type is indicative of a fault.
\cite{8806893} studies the adherence of an autonomous driving framework to ISO 26262 software standard. In~\cite{10.1007/978-3-030-26601-1_2} authors propose a graceful degradation design process to improve the automated driving continuation rate by analyzing tasks in the order of system-level, ECU-level, and microcontroller-level degradation.

There is a noticeable lack of a comprehensive fault modes, errors, and criticality analysis for a real fully-autonomous autonomous vehicle software implementation. We believe our paper addresses this gap, by presenting sensor level, algorithmic level, and system level faults and mitigating them.

%% file: sections/setup.tex
%The architecture of our safety-oriented software stack can be divided into five modules.
%The control module is responsible for controlling the vehicle.
%The localization module provides the location of the vehicle on the race track.
%The communication module relays data from the sensors and send control signals to the actuators, and is %responsible for maintaining communication with both the base station operators and race control.
%The perception module estimates the position of obstacles.
%Finally, the safety module ensures that the other modules behave in a manner that preserves the safety of the %vehicle on the track.

% The architecture required to drive the AV-21 race car is different from that needed for an on-road vehicle.
% This is due to the rules of the racing competition that have been placed on the race car.
% For this paper, we will use the racing rules defined for the most recent autonomous competition, which took place at the Las Vegas Motor Speedway (LVMS).

The Indy Autonomous Challenge features nine university teams, each using an AV-21 race car, which is an Indy lights car modified to drive autonomously. The AV-21 race car is retrofitted with a set of sensors and actuators: four RADAR sensors, six cameras and three LiDAR sensors. Each of the sensor modalities covers a 360-degree field-of-view around the vehicle. 
Furthermore, the vehicle is equipped with dual high-precision Global Navigation Satellite System (GNSS) receivers. 
This allows the vehicle to reach speeds of up to 175 mph around the oval race circuits - Indianapolis Motor Speedway (IMS) and the Las Vegas Motor Speedway (LVMS).

\begin{figure*}
    \centering
    \includegraphics[width=\linewidth]{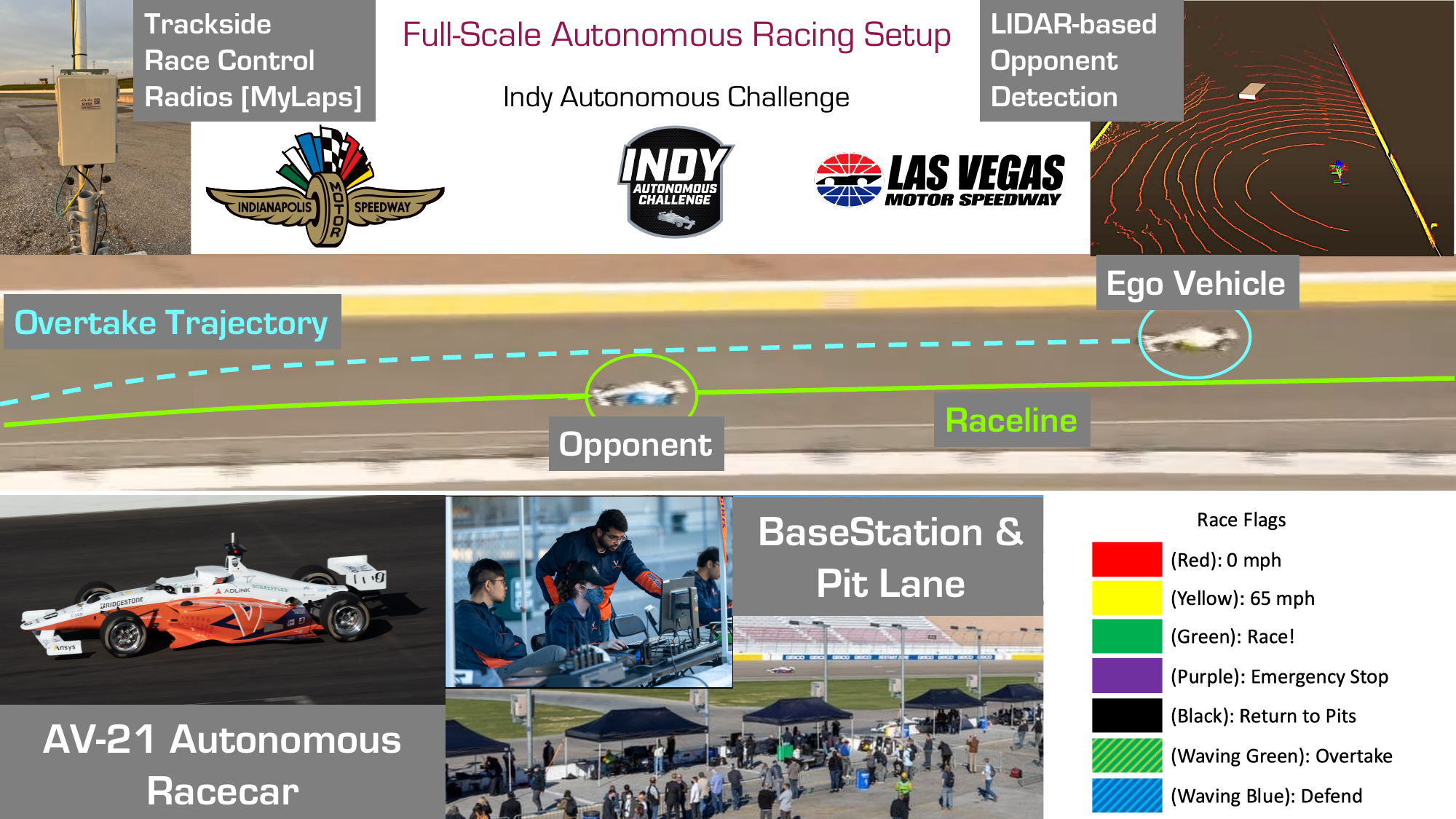}
    \caption{An overview of the autonomous racing setup in the Indy Autonomous Challenge held at the Indianapolis Motor Speedway and the Las Vegas Motor Speedway. The top-left gives an example of the radios used for communication between race control, the base station, and the vehicles on track. The top-right shows a LiDAR point cloud used to detect obstacles on track. In the center is an overtake trajectory, as performed during head-to-head racing. The bottom-left is an image of an AV-21 Autonomous Racecar used in the Indy Autonomous Challenge. The bottom-central shows the typical base station setup in pit lane. The bottom-right shows the different race flags that may be given to a vehicle by race control.}
    % \caption{An overview of the autonomous racing setup in the Indy Autonomous Challenge held at the Indianapolis Motor Speedway and the Las Vegas Motor Speedway. We implemented the HALO safety architecture on a full-scale fully autonomous AV-21 race car.}
    \label{fig:race_overview}
\end{figure*}

Figure~\ref{fig:race_overview} shows an overview of the autonomous racing setup and systems required for the race. It also outlines a few definitions that we will use throughout this paper. 
There are two competition modes - single-vehicle time-trials, and head-to-head racing. For the time-trial format, the goal is to go as fast as possible and put up the fastest lap time. For the head-to-head competition, the rules dictate to demonstrate the ability to overtake an opponent at a given speed, which incrementally increases in each lap.
%However, if an opponent vehicle is blocking the ego vehicle's race trajectory, it must also be capable of overtaking the opponent using an overtake trajectory.
For overtaking, the Ego race car must maintain a longitudinal separation from the Opponent race car. 
This required safe distance is 30 meters for the competition.

At the track, there are radio systems installed in place to help conduct the race and aid with the overall safety of the competition. An example of one of these radios is shown in the top-left of Figure~\ref{fig:race_overview}.
The racetrack is equipped with a network of mesh radios that enable teams to connect and communicate with the race car from their Pit Box, while the car is on the track.
% Teams have a Base Station computer in their pits which is able to obtain real-time telemetry data from the car, as well as issue safety-critical commands, in case of an emergency.
Teams have a Base Station computer in their pits, typically set up as in the bottom-central image of Figure~\ref{fig:race_overview}.
This base station is able to obtain real-time telemetry data from the car, as well as issue safety-critical commands, in case of an emergency.
Teams are forbidden to send any active control commands during the race. The race car is running autonomously \textit{on-its-own}. 
%Additionally, each vehicle must maintain communication with its base station at all times, so that a base station operator can intervene in an emergency.

%Another set of rules placed upon the race car are the requirements codified in the race flags.
A very important part of the race is adherence to Race Flags - which are relayed to the race car as radio signals over the mesh network. 
The race flags dictate constraints on the actions the race car can perform and also help communicate track status to the vehicle. This race control system is called MyLaps and it is also responsible for measuring the lap time and vehicle speeds during the autonomous race.
%These race flags are sent wirelessly by race control using the MyLaps system, resulting in a requirement for each vehicle to maintain communication with race control at all times in order to keep receiving these flags.
%The race flags used during the LVMS competition can be seen in Figure \ref{fig:race_overview}, along with the requirement it defines.
Depicted in the bottom-left of Figure~\ref{fig:race_overview} are the definitions of all the race flags.
The Red flag indicates that the vehicle must come to an immediate stop.
The Yellow flag implies caution, and directs the vehicle to remain below 65 mph on the track, or below 35 mph in the pit lane.
The Black flag means that the vehicle must return back to the pits from the track.
A Purple flag is issued to shut off the engine of the race car. This is used during an emergency or a crash/spin to prevent damage to the vehicle. 
Teams' software must respond to any flags issued by race control at all times, and must obey the flag constraints. 
The waving green and waving blue flags are used during overtaking maneuvers and indicate who is the attacker and the defender. 

%% file: sections/overview.tex
The goal of the HALO safety architecture is to prevent crashes.
It does this by leveraging existing fault tolerance methods to monitor our race car and mitigate any faults detected.
These faults can originate from code failure, where the code fails to execute; sensor failure, where a sensor provides bad or no data; or algorithmic failure, where the code produces an unsafe output. More discussion on failure modes can be found in Section~\ref{sec:safety}.
Our system aims to mitigate all of these possible faults in a holistic manner.

Before describing the HALO architecture that allows our autonomous race car to drive safely, we will briefly outline the components of the software stack that are not safety-related to provide context for what faults our safety architecture must account for.
There are many different ways to implement the software required to drive an race car autonomously.
For this paper we will describe the software that was used on our particular race car, but we acknowledge that there are other methods that may be used to achieve similar results.
Our autonomous race car's software has four safety-critical modules.
Each of these modules contains multiple nodes written in the ROS2 Foxy language, which communicates over a standard Data Distribution Service (DDS) middleware.
The four modules are:
    \begin{enumerate}
  \item Control module
  \item Localization module
  \item Communication module
  \item Perception module
\end{enumerate}

\begin{figure}
     \centering
     \includegraphics[width=\columnwidth]{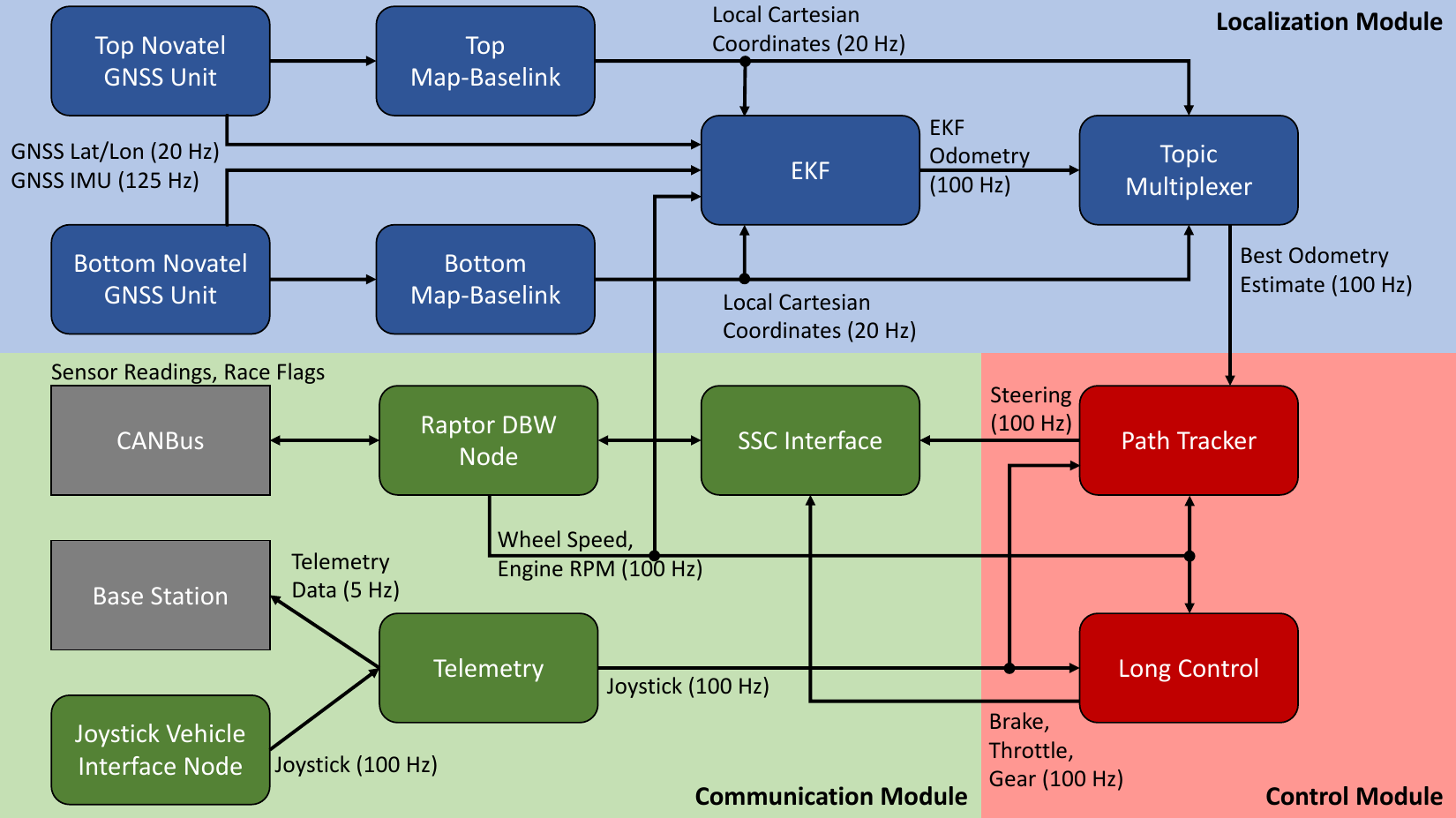}
     \caption{An overview of the software stack and the flow of data between nodes. The Control Module, shown in red, uses the wheel speed, engine RPM, and vehicle localization to generate actuator commands to drive the vehicle. The localization module, shown in blue, fuses several localization sources to generate a single best position estimate for the vehicle. The communication module, shown in green, is responsible for communicating with the CANBus and base station.}
     \label{fig:all_modules}
\end{figure}

Together, these modules control every aspect of the vehicle in a fully-autonomous manner.
The control module is responsible for controlling the steering, acceleration, and gearbox of the race car.
It takes as inputs the wheel speed, engine RPM, and vehicle localization, and generates accelerator and brake commands to track a set velocity, and steering angles to track a set path.
The localization module's goal is to provide an accurate estimate of the vehicle's position on the race track.
It fuses position data from two Global Navigation Satellite System (GNSS) units with Inertial Measurement Unit (IMU) data and Wheel Speed to determine the best estimate of the vehicle's position.
The communication module provides data from the low level hardware (Raptor Drive-By-Wire (DBW) system) to the software stack, and sends actuator commands from the vehicle's software stack to the low level hardware.
It is also responsible for maintaining communication with the base station in the pits and race control.
The perception module analyzes LiDAR point clouds to detect and track obstacles that must be avoided.
In Sections~\ref{sec:control},~\ref{sec:local},~\ref{sec:comm},~\ref{sec:perception}, we present a deep-dive of each of the modules with a focus on the safety requirements of each, identifying the failure modes, and the criticality of failures.

The nodes comprising these modules are critical components of the vehicle's operation.
The purpose of HALO is to maximize the safety of the vehicle by ensuring the proper behavior of these nodes.
This involves monitoring the activities of each node individually, and also their holistic interaction.
Section~\ref{sec:safety} presents the safety archetypes and overall fault analysis, followed by a description of the HALO safety architecture in Section~\ref{sec:halo}.

%% file: sections/control_module.tex
\noindent \textbf{Role:}
The control module is responsible for controlling the steering, acceleration, and gearbox of the AV-21 race car as well as to follow the race (flag) conditions at all times.
Shown in red in Figure \ref{fig:all_modules}, the control module consists of two nodes: Longitudinal (Long) Control and Lateral Control (Path Tracker).

%  \begin{figure}
%      \centering
%      \includegraphics[width=\columnwidth]{figs/control_module.pdf}
%      \caption{The control module, shown in red, uses the wheel speed, engine RPM, and vehicle localization to generate actuator commands to drive the vehicle.}
%      \label{fig:control_module}
%  \end{figure}

 % \begin{figure}
 %     \begin{subfigure}{0.49\textwidth}
 %         \centering
 %         \includegraphics[width=\columnwidth]{figs/control_module.pdf}
 %         \caption{The control module, shown in red, uses the wheel speed, engine RPM, and vehicle localization to generate actuator commands to drive the vehicle.}
 %         \label{fig:control_module}
 %     \end{subfigure}
 %     \hfill
 %     \begin{subfigure}{0.49\textwidth}
 %         \centering
 %         \includegraphics[width=\columnwidth]{figs/localization_module.pdf}       \caption{The localization module, shown in blue, fuses several localization sources to generate a single best position estimate for the vehicle.}
 %         \label{fig:localization_module}
 %     \end{subfigure}
 %     \hfill
 %     \begin{subfigure}{0.49\textwidth}
 %        \centering
 %        \includegraphics[width=\columnwidth]{figs/communication_module.pdf}  
 %        \caption{The communication module, shown in green, is responsible for communicating with the CANBus, race control, and base station.}
 %        \label{fig:communication_module}
 %     \end{subfigure}
 %     \caption{The data flow for each of the control, localization, and communication modules}
 % \end{figure}

\noindent \textbf{Inputs and Outputs:}
As shown in Table~\ref{table:control_module}, the control module takes the following signals as input:
\begin{enumerate}
    \item Wheel Speed: The \verb|Wheel Speed| measurement is reported on the CANBus by the Raptor Drive-By-Wire system
    \item Engine RPM: The engine RPM is contained as a field within \verb|Engine Report|
    \item Vehicle Localization: \texttt{Best Odometry}
    \item Joystick Commands: \texttt{Joystick}
\end{enumerate}
The control module uses these inputs to determine what speed the car should go at, what the steering angle should be, and what gear the vehicle needs to be in.
As output, the control module produces actuator commands for the accelerator, brake, steering column, and gear box:
\begin{enumerate}
    \item Accelerator actuator command: \verb|Accelerator Command|
    \item Brake actuator command: \verb|Brake Command|
    \item Steering column actuator command: \verb|Steering Command|
    \item Gear shifting command: \texttt{Gear Command}
\end{enumerate}

\noindent \textbf{Safety Concerns:}
The primary safety concern of the control module is to prevent the vehicle from crashing.
Examples of such unsafe behavior is attempting to turn too sharply or braking too aggressively at high speeds.
We next describe the operation of the all the nodes within the control module.
%The primary safety concern for the control module is having an emergency override from the base station available at all times.
%This override allows the base station operator to force the car to ignore the commands generated by the control module and instead use the joystick commands.

\begin{table*}[h]
\resizebox{\textwidth}{!}{\begin{tabular}{c|c|c|c|c}
    \textbf{Node} & \textbf{Data} & \textbf{I/O Type} & \textbf{Failure Type} & \textbf{HALO Safety Node} \\
    \hline
    Long Control & Joystick & Input & Data Health & Graceful Stop \\
     & Graceful Stop Signal & Input & Node Health & Node Health Monitor \\
     & Wheel Speed & Input & Data Health & None \\
     & Engine Report & Input & Data Health & None \\
     & Long Control Heartbeat & Output & Node Health & Node Health Monitor \\
     & Accelerator Command & Output & Data Health & SSC Interface \\
     & Brake Command & Output & Data Health & SSC Interface \\
     & Gear Command & Output & Data Health & SSC Interface \\
    \hline
    Path Tracker & Joystick & Input & Data Health & Graceful Stop \\
     & Desired Path & Input & Behavioral-Safety & Path Tracker \\
     & Best Odometry & Input & Data Health & Graceful Stop \\
     & Wheel Speed & Input & Data Health & None \\
     & Path Tracker Heartbeat & Output & Node Health & Node Health Monitor \\
     & Steering Command & Output & Data Health & SSC Interface \\
    \hline
\end{tabular}}
\caption{\label{table:control_module}Control Module inputs and outputs. For each signal we describe what the data is, and whether it is an input to the node or output from the node. We also note for each signal what the potential failure type for that data is, and which HALO Safety Node (if any) is responsible for handling that failure. The data flow into and out of the Control Module can be tracked into nodes, shown in Tables~\ref{table:localization_module},~\ref{table:communication_module}}
\end{table*}

\subsection{Longitudinal Control}
\label{sec:long_control}
The Long Control Node is responsible for controlling the vehicle's accelerator, brake, and gearbox to maintain a desired velocity. This is referred to as longitudinal control for an autonomous vehicle.
%This node takes in the velocity of the wheels, the engine rpm, and the accelerator and brake commands from the base station joystick as its inputs.
%The desired velocity is a parameter that can be set by publishing a float value on the \texttt{/vehicle/set\_desired\_velocity} ROS topic.

Using the vehicle's current velocity derived from the wheel speed, the Long Control Node uses two PID controllers to bring the vehicle to the desired velocity.
One PID controller controls the accelerator and the other controls the brake.
%The outputs of these controllers are sent as float values from 0 to 100, representing the percent accelerator and brake requested, and sent to the Steering and Speed Control (SSC) Interface, which will pass the data to the Raptor Drive-By-Wire system.
The outputs of these controllers represent the amount that the accelerator and brake should be applied, and are sent to the Steering and Speed Control (SSC) Interface, which will pass the data to the Raptor Drive-By-Wire system.
In addition to accelerator and brake commands, the Long Control Node also controls the gearbox.
%The node has vehicle velocity thresholds and engine RPM thresholds defined for both shifting up in gear and shifting down in gear.
%If the vehicle's velocity and engine RPM both exceed the threshold for the next gear, the node will request the vehicle shift up one gear.
%Conversely, if the vehicle's velocity and engine RPM both drop below the threshold for the previously gear, the node will request the vehicle shift down one gear.
Gear shifting is calculated based on the engine RPM and the velocity of the race car, and sent as an integer value to the SSC Interface.

\noindent \textbf{Safety Concerns:}
The Long Control Node has several in-built safety features.
The first safety feature is a check that prevents the node from publishing an accelerator and a brake command simultaneously. This prevents drivetrain damage by trying to apply both the brake and the accelerator at the same time.
The second safety feature is a check on the data limits of the accelerator and brake. If either of these commands exceed their acceptable range the node reduces the value to the maximum allowed value. Similarly, if either of these commands fall below zero the node will set the value to zero.
The third safety feature is an emergency override from the base station. 
During testing on the track, the joystick operator in the pit lane can override the accelerator and braking commands. 
%The node listens to the commands coming from the joystick, and overrides the autonomous controller if input from the joystick is detected.
%In the case of the brake, if the base station operator applies the brake the node will publish the brake value coming from the joystick instead of its own brake or accelerator values.
%In the case of the accelerator, the node sets its maximum acceptable value for the throttle to be the accelerator value coming from the joystick.
Additionally, the base station operator can make an emergency stop by pressing two bumpers on the joystick simultaneously. This causes an immediate engine shutdown on the vehicle.

\subsection{Lateral Control - Path Tracker}
\label{sec:path_tracker}
The Path Tracker Node is responsible for controlling the vehicle's steering column to follow a reference path. 
A path is defined using a sequence of waypoints in a local north-east-down (NED) coordinate frame.

Path Tracker node takes as input the wheel speed, the vehicle's position, and the joystick commands from the base station.
The path for the node to follow is an internal state of this node. %, and can be set by publishing a string value on the \texttt{/switch\_path} topic.
The Path Tracker node uses three paths - ``raceline'', which takes the inner line throughout the race track; ``overtake,'' which takes the outer line of the race track for the purposes of overtaking an opponent; and ``pits,'' which enters the pit lane from the inner line of the race track, goes through the pits, and then merges back onto the inner line.
%The paths are defined in files that are read when the node is started.
Each path consists of equally spaced waypoints defined in a local NED coordinate frame relative to a shared origin point.

The Path Tracker Node uses an Ackermann-adjusted pure-pursuit algorithm \cite{pure_pursuit} to follow the chosen path.
A Model Predictive Control \cite{mpc} approach could also be used in lieu of the pure-pursuit algorithm.
The node finds the waypoint on the path that is closest laterally to the current position of the vehicle.
From this waypoint, the node then uses a dynamic lookahead distance to find a waypoint ahead of the vehicle.
The lookahead distance is computed as a function of the vehicle's velocity. Up to 35 mph, the lookahead distance is set at 15 meters. From 35 mph to 100 mph, the lookahead distance increases linearly from 15 meters to 50 meters. 
Over 100 mph the lookahead distance is capped at 50 meters.
%The waypoint chosen from this lookahead distance is then used by the pure-pursuit algorithm to determine the steering angle, which is published as a float value from -24.0 to 24.0, representing the angle of the steering column, and is sent to the SSC Interface.

\noindent \textbf{Safety Concerns:}
The Path Tracker Node has several safety features built into it.
The first safety feature is a check on the data limits of the steering angle. If the command exceeds the maximum angle in either direction, the node reduces the value to be within the acceptable range.
The second safety feature is a dynamic adjustment of the maximum steering angle, calculated as a function of the vehicle's velocity. Below 35 mph the vehicle's steering column has a range from 24 degrees in either direction. From 35 mph to 100 mph, the steering range decreases linearly from 24 degrees to 10 degrees. Over 100 mph the steering column has a set range of 10 degrees in either direction. This dynamic adjustment of the maximum steering angle prevents the vehicle from attempting to turn too sharply at high speeds, which could result in losing control of the vehicle.
The third safety feature is to check if a valid path is being published at all time. Especially while switching to a new path, the node verifies that the requested path exists.

%% file: sections/localization_module.tex
\noindent \textbf{Role:}
The goal of the localization module is to provide an accurate estimate of the vehicle's position on the race track.
Shown in blue in Figure \ref{fig:all_modules}, the localization module consists of six nodes: Top GNSS, Bottom GNSS, Top Map-Baselink, Bottom Map-Baselink, EKF, and Topic Multiplexer.

\noindent \textbf{Inputs and Outputs:}
the localization module takes the following signals as input (also listed in Table~\ref{table:localization_module}):
\begin{enumerate}
    \item Global Navigation Satellite System (GNSS) data from the two Novatel units in the form of latitude, longitude, latitude standard deviation, and longitude standard deviation, published at a rate of 20 Hz: \verb|Top GNSS Lat/Lon| and \verb|Bottom GNSS Lat/Lon|
    \item Inertial Measurement Unit (IMU) data from the two Novatel units, published at a rate of 125 Hz: \verb|Top GNSS IMU| and \verb|Bottom GNSS IMU|
    \item Wheel Speed: \verb|Wheel Speed|
\end{enumerate}
It uses these inputs to determine the location of the vehicle in a local Cartesian (x,y) coordinate frame, published on the topic \texttt{Best Odometry}.

\noindent \textbf{Safety Concerns:}
The safety concerns of the localization module are the data rate and data accuracy.
% During a race, the vehicle is moving very fast.
% If the localization data is published too slowly then the vehicle's controls will also update slowly, which could cause the vehicle to either miss a turn or overcorrect leading to a crash.
Data rate is a concern because when driving at 150 mph, the vehicle will travel over 10 feet in between successive GNSS readings, essentially driving blindly.
%\hl{example of at 150mph, at 20 Hz the car has travelled XX feet - blindly - risky}
Low accuracy data is also a concern, because inaccurate localization could cause the vehicle to crash.
We next discuss the localization methodology for the vehicle.

\begin{table*}[h]
\resizebox{\textwidth}{!}{\begin{tabular}{c|c|c|c|c}
    \textbf{Node} & \textbf{Data} & \textbf{I/O Type} & \textbf{Failure Type} & \textbf{HALO Safety Node} \\
    \hline
    GNSS Top & Top GNSS Lat/Lon & Output & Raw Data & Graceful Stop \\
     & Top GNSS IMU & Output & Raw Data & None \\
    \hline
    GNSS Bottom & Bottom GNSS Lat/Lon & Output & Raw Data & Graceful Stop \\
     & Bottom GNSS IMU & Output & Raw Data & None \\
    \hline
    Map-Baselink Top & Top GNSS Lat/Lon & Input & Raw Data & Graceful Stop \\
     & Top Cartesian Coordinates & Output & Data Health & Topic Multiplexer \\
    \hline
    Map-Baselink Bottom & Bottom GNSS Lat/Lon & Input & Raw Data & Graceful Stop \\
     & Bottom Cartesian Coordinates & Output & Data Health & Topic Multiplexer \\
    \hline
    Convert IMU Top & Top GNSS IMU & Input & Raw Data & None \\
     & Top GNSS IMU Corrected & Output & Data Health & None \\
    \hline
    Convert IMU Bottom & Bottom GNSS IMU & Input & Raw Data & None \\
     & Bottom GNSS IMU Corrected & Output & Data Health & None \\
    \hline
    EKF & Top Cartesian Coordinates & Input & Data Health & Topic Multiplexer \\
     & Top GNSS IMU Corrected & Input & Data Health & None \\
     & Bottom Cartesian Coordinates & Input & Data Health & Topic Multiplexer \\
     & Bottom GNSS IMU Corrected & Input & Data Health & None \\
     & Wheel Speed & Input & Raw Data & None \\
     & EKF Odometry & Output & Data Health & Topic Multiplexer \\
    \hline
    Topic Multiplexer & Top GNSS Odometry & Input & Data Health & Topic Multiplexer \\
     & Bottom GNSS Odometry & Input & Data Health & Topic Multiplexer \\
     & EKF Odometry & Input & Data Health & Topic Multiplexer \\
     & MyLaps Race Flags & Input & Data Health & Topic Multiplexer \\
     & Spoofed Race Flags & Input & Data Health & Topic Multiplexer \\
     & Wheel Speed & Input & Raw Data & None \\
     & Best Odometry & Output & Data Health & Graceful Stop \\
     & No Odometry Signal & Output & Data Health & Graceful Stop \\
     & Best Race Flags & Output & Data Health & Topic Multiplexer \\
     & Topic Multiplexer Heartbeat & Output & Node Health & Node Health Monitor \\
     \hline
\end{tabular}}
\caption{\label{table:localization_module}Localization Module inputs and outputs. For each signal we note a description of the data is, whether it is an input to the node or output from the node, the signal's potential failure type, and which HALO Safety Node (if any) is responsible for handling that failure. The data flow into and out of the Localization Module can be tracked into nodes, shown in Tables~\ref{table:control_module},~\ref{table:communication_module}}
\end{table*}

%  \begin{figure}
%      \centering
%      \includegraphics[width=\columnwidth]{figs/localization_module.pdf}
%      \caption{The localization module, shown in blue, fuses several localization sources to generate a single best position estimate for the vehicle.}
%      \label{fig:localization_module}
%  \end{figure}

\subsection{Localization Methodology - Sensor Fusion}
\noindent \textbf{Map-Baselink} The Map-Baselink Node takes the latitude and longitude values reported by the GNSS unit and converts them into an x and y position in a local north-east-down (NED) coordinate.
This coordinate transformation is done using a an arbitrary origin and using a Geodesic of the Earth.
In addition to the local (x,y) position, the Map-Baselink Node also computes $\theta$, the heading of the vehicle using hysteresis. The node computes the change in position from one GNSS measurement to the next, and reports that angle as the heading of the vehicle.
There is a separate Top Map-Baselink Node and Bottom Map-Baselink Node for the top and bottom GNSS units, reporting the \texttt{Top Cartesian Coordinates} and \texttt{Bottom Cartesian Coordinates} respectively.
%These are the basic localization topics used by the vehicle.

% The Map-Baselink Node takes the latitude and longitude values reported by the GNSS units and converts them into an x and y position in a local north-east-down (NED) coordinate frame. %\hl{We convert geodetic coordinates specified by lat, lon, and h to the local north-east-down (NED)}. 
% %This conversion uses the same origin point as the path files read into the Path Tracker.
% In addition to the local (x,y) position, the Map-Baselink Node also computes $\theta$, the heading of the vehicle using hysteresis. The node computes the change in position from one GNSS measurement to the next, and reports that angle as the heading of the vehicle.
% There is a separate Top Map-Baselink Node and Bottom Map-Baselink Node for the top and bottom GNSS units, and they publish on the topics \texttt{/novatel\_top/dyntf\_odom} and \texttt{/novatel\_bottom/dyntf\_odom} respectively.
%These are the basic localization topics used by the vehicle.

%\subsection{Extended Kalman Filter (EKF)}
\label{sec:ekf}
\noindent \textbf{Extended Kalman Filter} The GNSS localization process occurs at the slower rate of 20 Hz. 
%When traveling at racing speeds upwards of 100 mph, the GNSS data is not being reported often enough to keep the vehicle safely on the track. 
The GNSS positional data is fused with linear acceleration and angular velocity measurements using an Extended Kalman Filter (EKF)~\cite{ekf}. 
The EKF takes the 20 Hz Cartesian coordinates from the Top and Bottom Map-Baselink and fuses them with 125 Hz Inertial Measurement Unit (IMU) data from the GNSS units and the 100 Hz Wheel Speed data. From these sources, the EKF creates a single estimate of the vehicle's position at 100 Hz.

\subsection{Localization Redundancy - Topic Multiplexer}
The localization process can produce position estimates that do not accurately reflect the position of the vehicle. 
Additionally, one GNSS unit can be more accurate than the other. 
It is the job of the Topic Multiplexer Node to determine which source of localization is the most accurate, and report those values as the position of the vehicle, \texttt{Best Odometry}.
When determining which source of localization to use, the Topic Multiplexer Node always prefers the 100 Hz EKF localization over the 20 Hz GNSS localization.
The node monitors the covariance of the EKF localization estimation and, if the covariance is greater than an acceptable threshold, the 100 HZ data is considered imprecise and the topic multiplexer defaults to reporting the slower 20 Hz GNSS localization.
When reporting GNSS localization, the topic multiplexer prioritizes the higher accuracy output between the two GNSS systems. Slowing down to a slower (but accurate) GNSS is better than inaccurate (but faster) EKF estimates. When this switch in the source occurs, the race car also slows down appropriately since the localization data frequency has reduced by a factor of 5. 
%In this way, we can ensure that the vehicle is always receiving the best possible source of localization.

%% file: sections/communication_module.tex
\noindent \textbf{Role:}
The goal of the communication module is to provide sensor readings from the low level hardware (Raptor Drive-By-Wire (DBW) system) to the vehicle's software stack and send actuator commands from the software down to the low level hardware. 
In addition, the race car needs to maintain a radio link with the base station in the Pits and with with the MyLaps race control flags system. 
Shown in green in Figure \ref{fig:all_modules}, the communication module consists of five nodes: Raptor DBW, Steering and Speed Control (SSC) Interface, Joystick Vehicle Interface, Race Flag Input, and Telemetry.

\begin{table*}[h]
\resizebox{\textwidth}{!}{\begin{tabular}{c|c|c|c|c}
    \textbf{Node} & \textbf{Data} & \textbf{I/O Type} & \textbf{Failure Type} & \textbf{HALO Safety Node} \\
    \hline
    SSC Interface & Best Odometry & Input & Data Health & Graceful Stop \\
     & Joystick Emergency Shutdown & Input & Data Health & None \\
     & Base Station Heartbeat & Input & Data Health & Graceful Stop \\
     & Diagnostics Heartbeat & Input & Node Health & Graceful Stop \\
     & Accelerator Command & Input & Data Health & SSC Interface \\
     & Brake Command & Input & Data Health & SSC Interface \\
     & Gear Command & Input & Data Health & SSC Interface \\
     & Steering Command & Input & Data Health & SSC Interface \\
     & Raptor System Report & Input & Data Health & None \\
     & Wheel Speed & Input & Data Health & None \\
     & Raptor Accelerator Command & Output & Data Health & SSC Interface \\
     & Raptor Brake Command & Output & Data Health & SSC Interface \\
     & Raptor Gear Command & Output & Data Health & SSC Interface \\
     & Raptor Steering Command & Output & Data Health & SSC Interface \\
    \hline
    Raptor DBW Node & Raptor Pedal Command & Input & Data Health & SSC Interface \\
     & Raptor Brake Command & Input & Data Health & SSC Interface \\
     & Raptor Gear Command & Input & Data Health & SSC Interface \\
     & Raptor Steering Command & Input & Data Health & SSC Interface \\
     & Wheel Speed & Output & Raw Data & None \\
     & MyLaps Race Flags & Output & Data Health & Topic Multiplexer \\
     & Engine Report & Output & Raw Data & None \\
     & Raptor System Report & Output & Raw Data & None \\
     & Raptor Diagnostics Report & Output & Raw Data & Graceful Stop \\
    \hline
    Telemetry & Basestation Joystick & Input & Data Health & None \\
     & Joystick & Output & Data Health & Graceful Stop \\
     & Telemetry Data & Output & Data Health & None \\
\end{tabular}}
\caption{\label{table:communication_module}Communication Module inputs and outputs. For each signal we note a description of the data is, whether it is an input to the node or output from the node, the signal's potential failure type, and which HALO Safety Node (if any) is responsible for handling that failure. The data flow into and out of the Control Module can be tracked into nodes, shown in Tables~\ref{table:control_module},~\ref{table:localization_module}}
\end{table*}

\noindent \textbf{Inputs and Outputs:}
The communication module takes the following signals as input (also listed in Table~\ref{table:communication_module}):
\begin{enumerate}
    \item Vehicle ECU data from the CANBus
    \item Race flags from the CANBus
    \item Joystick commands from the base station
    %\item Spoofed race flags from the base station
    %    \item Accelerator actuator command:\verb|/joystick|\\\verb|/accelerator_cmd|
    \item Actuator commands for the Raptor DBW system: \verb|Accelerator Command|, \verb|Brake Command|, \\\verb|Steering Command|, and \verb|Gear Command|
\end{enumerate}
The communication module provides two outputs. First, it converts the actuator commands into bytes to be put on the CANBus for the Raptor DBW system. Second, it takes the inputs from the CANBus and the base station, and converts them into ROS2 topics for the other modules to use:
\begin{enumerate}
    \item Wheel Speed: \verb|Wheel Speed|
    \item Engine RPM: \verb|Engine Report|
    \item Race flags: \verb|MyLaps Race Flags|
    \item Joystick commands: \verb|Raw Joystick|
    %\item Spoofed race flags: \verb|/raptor_dbw_interface|\\\verb|/rc_to_ct_bs|
\end{enumerate}
%As inputs, the communication module takes data from the CANBus, the base station, and the control module.
%From the CANBus, the inputs are sensor data and race flags.
%From the base station, the inputs are joystick commands and spoofed race flags.
%From the control module, the inputs are the actuator commands for the low-level hardware
%As outputs, the communication module provides the current race flag, the joystick commands, and the sensor data to the control module.
%The communication module also provides actuator commands as outputs, published on the CANBus.

\noindent \textbf{Safety Concerns:}
The primary safety concern of the communication module is a loss of communication.
A loss of communication with the CANBus means no sensor data for the software stack, but more critically it also implies that the control module cannot send commands to the Raptor DBW system to control/actuate the vehicle.
A loss of communication with the base station presents a danger to the vehicle because it removes the base station operator's ability to intervene in the case of an emergency.
Similarly a loss of communication with the MyLaps race control prevents the vehicle from being directed autonomously through flag signaling.
%We next discuss the communication nodes and the data they process.

%  \begin{figure}
%      \centering
%      \includegraphics[width=\columnwidth]{figs/communication_module.pdf}
%      \caption{The communication module, shown in green, is responsible for communicating with the CANBus, race control, and base station}
%      \label{fig:communication_module}
%  \end{figure}

\subsection{Communication With Sensors}
\noindent \textbf{Raptor DBW Node} The Raptor DBW Node in the communication module is responsible for sending control commands from the control module to the actuators. The Raptor DBW Node subscribes to the control commands coming from the SSC Interface, and converts them into binary vectors to be published on the CANBus.
%The communication module is to communicate with the sensors of the vehicle. 
The Raptor DBW Node also provides sensor data off of the CANBus, and converts it into the a ROS2 message.
This data includes information about the vehicle's engine published in the \texttt{Engine Report}, information about the wheel speed published in \texttt{Wheel Speed}, and information regarding diagnostics of the low-level control system published in \texttt{Raptor Diagnostics Report}.
The Raptor DBW Node also listens for \texttt{MyLaps Race Flags} coming from race control. The data from race control is sent to a radio that publishes the data onto the CANBus. The Raptor DBW Node converts this data into a ROS2 message.
%Values for the race flag data can be seen in Figure \ref{fig:race_overview}, and is published on the topic $/raptor\_dbw\_interface/rc\_to\_ct$.
 % TODO: Clear up the confusion between this and what you say in the Control Module

\subsection{Communication With Base Station}
\noindent \textbf{Telemetry} The Telemetry node is responsible for sending data from the vehicle to the base station using the mesh radio network.
%The Telemetry node listens to data on the vehicle that is relevant to the base station operator, and republishes it on a series of topics to be communicated to the base station.
The data sent to the base station includes the engine temperature, the localization accuracy, the vehicle's current and desired velocity, the lateral error from the vehicle to the set path, the race flag, the vehicle's position, and the reason for the most recent emergency stop (if any).
This data is published at 5 Hz to reduce the network bandwidth between the vehicle and the base station.

\noindent \textbf{Joystick Vehicle Interface Node} The Joystick Vehicle Interface Node is responsible for sending joystick commands from the base station to the vehicle using the mesh radio network. This node listens for commands coming from the joystick attached to the base station, and relays them to the vehicle as \texttt{Raw Joystick} at a rate of 100 Hz.
%The Race Flag Input Node is another base station node, and is responsible for sending spoofed race flags from the base station to the vehicle on the topic $/raptor\_dbw\_interface/rc\_to\_ct\_bs$.

\noindent \textbf{Safety Concerns:}
%The main safety concern of the communication module is the loss of communication with either the CANBus or the base station.
To address a loss of communication with the CANBus, the Raptor DBW Node produces a heartbeat to test communications with the sensors and actuators. If the low-level hardware does not receive this heartbeat, the engine will shut off to prevent a loss of control of the vehicle.
To address a loss of communication with the base station, the joystick commands include a heartbeat. If the vehicle does not receive this heartbeat for some time, the vehicle will be brought to a stop. This is discussed in detail in Section \ref{sec:graceful_stop}.

%% file: sections/perception_module.tex
\noindent \textbf{Role:}
The goal of the perception module is to detect and track obstacles. These could be static (air-filled pylons) or dynamic (opponent race car).
%on track for the purposes of avoiding a crash.
%Shown in Figure \ref{fig:perception_module}, 
Our software uses LiDAR and RADAR data for perception. The race car contains six cameras as well and these could also be integrated into the pipeline. 
%The LiDAR-based perception consists of five nodes: Point Cloud Filtering, Ground Detection, Wall Detection, Euclidean Clustering, and Object Detection.
The perception module consists of seven nodes: Ground Detection, Point Cloud Downsample, Polygon Filter, Point Cloud Clustering, DNN Inference, RADAR Filtering, and EKF.
% The LiDAR-based perception starts by cleaning the data using Ground Detection and Point Cloud Filtering nodes.
% The clean data is then fed into two different inference nodes: Clustering Detections and DNN Detections.
% The RADAR-based perception uses a single node, Radar Detections, to detect obstacles from the raw RADAR data.
% All of these detections are then fed into an EKF

\begin{table*}[h]
\resizebox{\textwidth}{!}{\begin{tabular}{c|c|c|c|c}
    \textbf{Node} & \textbf{Data} & \textbf{I/O Type} & \textbf{Failure Type} & \textbf{HALO Safety Node} \\
    \hline
    % LiDAR Front & /luminar\_front\_points & Output & Raw Data & None \\
    % \hline
    % LiDAR Left & /luminar\_left\_points & Output & Raw Data & None \\
    % \hline
    % LiDAR Right & /luminar\_right\_points & Output & Raw Data & None \\
    % \hline
    % Point Cloud Transformer Front & /luminar\_front\_points & Input & Raw Data & None \\
    %  & /luminar\_front\_points/filtered & Output & Data Health & None \\
    % \hline
    % Point Cloud Transformer Left & /luminar\_left\_points & Input & Raw Data & None \\
    %  & /luminar\_left\_points/filtered & Output & Data Health & None \\
    % \hline
    % Point Cloud Transformer Right & /luminar\_right\_points & Input & Raw Data & None \\
    %  & /luminar\_right\_points/filtered & Output & Data Health & None \\
    % \hline
    % LiDAR Fusion & /luminar\_front\_points/filtered & Input & Data Health & None \\
    %  & /luminar\_left\_points/filtered & Input & Data Health & None \\
    %  & /luminar\_right\_points/filtered & Input & Data Health & None \\
    %  & /points\_fused & Output & Data Health & None \\
    % \hline
    LiDAR & Raw LiDAR Point & Output & Raw Data & None \\
    \hline
    Ray Ground Classifier & Raw LiDAR Points & Input & Data Health & None \\
     & Ground Points & Output & Data Health & None \\
     & Non-Ground Points & Output & Data Health & None \\
    \hline
    % Right Wall Detector & /novatel\_top/dyntf\_odom & Input & Data Health & Topic Multiplexer \\
    %  & /points\_ground & Input & Data Health & None \\
    %  & /points\_nonground & Input & Data Health & None \\
    %  & /real\_points & Output & Data Health & None \\
    %  & /right\_wall & Output & Data Health & None \\
    %  & /left\_wall & Output & Data Health & None \\
    % \hline
    % Euclidean Cluster & /real\_points & Input & Data Health & None \\
    %  & /lidar\_bounding\_boxes & Output & Data Health & None \\
    % \hline
    % Right Wall Points Pub & /right\_wall & Input & Data Health & None \\
    %  & /points\_right\_wall & Output & Data Health & None \\
    %  & /path\_right\_wall & Output & Data Health & None \\
    % \hline
    % Left Wall Points Pub & /left\_wall & Input & Data Health & None \\
    %  & /points\_left\_wall & Output & Data Health & None \\
    %  & /path\_left\_wall & Output & Data Health & None \\
    % \hline
    % Heading Estimator & /lidar\_bounding\_boxes & Input & Data Health & None \\
    %  & /novatel\_top/dyntf\_odom & Input & Data Health & Topic Multiplexer \\
    %  & /telemetry/bounding\_box\_array & Output & Data Health & None \\
    %  & /telemetry/bb\_distance & Output & Behavioral-Safety & SSC Interface \\
    % \hline
    Point Cloud Downsample & Non-Ground Points & Input & Data Health & None \\
     & Downsampled Points & Output & Data Health & None \\
    \hline
    Polygon Filter & Downsampled Points & Input & Data Health & None \\
     & Track-Filtered Points & Output & Data Health & None \\
    \hline
    Clustering Detections & Track-Filtered Points & Input & Data Health & None \\
     & Clustering Opponent Detections & Output & Behavioral-Safety & None \\
    \hline
    DNN Inference & Raw LiDAR Points & Input & Data Health & None \\
     & DNN Opponent Detections & Output & Behavioral-Safety & None \\
    \hline
    RADAR Front & Raw RADAR & Output & Raw Data & None \\
    \hline
    RADAR Filtering & Raw RADAR & Input & Raw Data & None \\
     & Radar Opponent Detections & Output & Behavioral-Safety & None \\
    \hline
    EKF & Clustering Opponent Detections & Input & Behavioral-Safety & None \\
     & DNN Opponent Detections & Input & Behavioral-Safety & None \\
     & Radar Opponent Detections & Input & Behavioral-Safety & None \\
     & Best Odometry & Input & Data Health & Graceful Stop \\
     & Target Vehicle Odometry & Output & Behavioral-Safety & SSC Interface \\
     & Target Vehicle Acceleration & Output & Behavioral-Safety & SSC Interface \\
    \hline
\end{tabular}}
\caption{\label{table:perception_module}Perception Module inputs and outputs. For each signal we note a description of the data is, whether it is an input to the node or output from the node, the signal's potential failure type, and which HALO Safety Node (if any) is responsible for handling that failure. The data flow can be tracked from raw LiDAR points to opponent detections.}
\end{table*}

\noindent \textbf{Inputs and Outputs:}
As shown in Table~\ref{table:perception_module}, the perception module takes LiDAR point clouds, RADAR tracks, and vehicle localization as inputs. 
The perception module provides the following outputs:
\begin{enumerate}
    \item The locations and headings of obstacles on the track: \verb|Target Vehicle Odometry|
    \item The acceleration of the obstacles: \verb|Target Vehicle Acceleration|
\end{enumerate}

\noindent \textbf{Safety Concerns:}
The primary safety concern of the perception module is the detection and mitigation of False Positives and False Negatives. 
A false positive is the perception stack seeing an obstacle that does not exist, and a false negative is not seeing an obstacle that is present.
False negatives present a highly critical failure mode as the vehicle might not act to avoid a collision with an obstacle, not detected ahead of time.
False positive detection can lead to erratic behavior as the vehicle attempts to avoid a non-existent obstacle.
We next discuss the perception pipeline in detail.

%  \begin{figure}
%      \centering
%      \includegraphics[width=0.8\columnwidth]{figs/perception_module.pdf}
%      \caption{The perception module, shown in purple, uses filtering and clustering algorithms to perform object detection on LiDAR point clouds.}
%      \label{fig:perception_module}
%  \end{figure}

\subsection{Filtering}
The perception module begins with raw point cloud data from each of the three LiDARs.
This point cloud data is automatically fused into a single point cloud at the sensor level.
This point cloud is a dense collections of points, so the first step in the pipeline is to filter the point clouds down to fewer points.
The Ray Ground Classifier Node uses ground plane detection \cite{plane_detection} to distinguish ground points and non-ground points. 
The Point Cloud Downsample Node then uses a voxel downsampling algorithm to reduce the number of non-ground points.
The final filter is the Polygon Filter Node, which uses predefined polygon areas to filter out points which are outside the race track. This allows the vehicle to ignore noise caused by grass on the interior of the track, and obstacles detected in the stands.
% The Filtering Node starts with range-based filtering to ignore points that are too far away to be relevant, or are too close and likely belong to the ego vehicle itself. The filtering node also uses angle-based filtering to remove points that are too high above the vehicle and unnecessary for driving on a flat race track.
% In addition to these filters, the perception module also uses ground plane detection \cite{plane_detection} to separate the points into ground points and non-ground points. 
%Since there are no features for our vehicle to detect on the ground, our perception pipeline filters out all ground points.
% The final filter used by the perception module detects the track boundaries, so that the vehicle can filter out all points that are outside of the race track. This allows the vehicle to ignore noise caused by grass on the interior of the track, and obstacles detected in the stands.

\subsection{Object Detection and Tracking}
At this point, the only remaining points are those that belong to obstacles on the track.
Euclidean clustering~\cite{euclidean_clustering} is performed on these points, to cluster them together into objects.
Any cluster containing a critical mass of points are detected as an object, and are fitted to a bounding box to get the size and orientation of the object.
To increase detection accuracy, we also perform object detection using a DNN for redundancy.
Since the DNN is trained on dense point clouds, we use the raw LiDAR data as its input.
%Objects can then be analyzed to determine their distance and heading.
% The distance to the obstacle is calculated as the distance from the baselink of the ego vehicle to the centroid of the bounding box.
%Because the clustering algorithm often returns points that are in slightly different positions, the bounding box usually has a different orientation in each point cloud. Therefore, instead of using the orientation of the bounding box to determine heading, a hysteresis method is used instead, comparing the change in position of the centroid of the bounding box from one point cloud to the next.
In addition to these two LiDAR-based detection methods, we also use a RADAR-based detection method, using the front-facing RADAR.

These three methods, Clustering, DNN, and RADAR, are then fused into a single best detection using the same EKF configuration as in Section~\ref{sec:ekf}.
This fusion is helpful because the detection methods are most effective at different ranges. For example, the RADAR can detect obstacles that are too distant to appear in the LiDAR data.
The EKF also allows us to estimate the obstacle's position between successive updates from the object detection nodes, allowing for greater awareness of the obstacle's position.

\noindent \textbf{Safety Concerns:}
As mentioned earlier, the primary safety concern for the perception module is false negative and false positive detection from the LiDARs.
After the Euclidean Clustering algorithm, only clusters containing a enough points to exceed a preset threshold are considered to be an object.
If this threshold is set too low, the perception pipeline will detect non-existent obstacles as being present, usually from ground points that get misclassified as non-ground.
If a non-existent obstacle is detected, the vehicle may swerve to avoid an obstacle that only appears for a fraction of a second.
Alternatively, if the object detection threshold is set too high, the perception pipeline may not see a vehicle on track if not enough points pass through the filter to the clustering algorithm.
In this case, the ego vehicle could make an early attempt to complete an overtake, thinking it has passed the opponent vehicle.
When a middle-ground threshold is chosen for obstacle detection, these problems are rarer but either can still occur.
Similarly, the DNN detection algorithm can also miss detections or see false detections.
As DNNs are black-box models, fine-tuning them to eliminate this behavior is impossible.
The RADAR detection method is also not error free.
Since these models can all produce false positives and false negatives during runtime, detecting them is imperative.
%, leading to the need for an algorithm that can handle momentary appearance of ghosts, or momentary disappearance of real obstacles.
%However, the perception pipeline does not concern itself with solving this problem. It merely reports the perception data as available, and allows another node take responsibility for safety when making a decision based on perception data.
In Section \ref{sec:halo} we will discuss a behavioral safety node that can handle this situation.

%% file: sections/safety_module.tex
%The goal of the safety module is to prevent damage to the ego vehicle, and protect other vehicles on track as much as possible.
The overall safety goal of the vehicle is to prevent any crashes.
%Within this goal, the safety module is also trying to maximize the amount of time that the vehicle can spend on track.
The safety requirements are partially codified in the form of the race flag rules defined in Section~\ref{sec:setup} that must be obeyed at all times.
In addition to the flag rules, there are other failure modes that pose a risk to the safe operation of the ego vehicle and, therefore, must be handled appropriately by our HALO safety framework. 
A piece of code crashing means losing some functionality on the vehicle, and presents danger based on the severity of the code failure.
For instance, losing the ability to precisely localize, whether because of an inaccurate localization estimate, or because of a lack of localization data, could lead to a catastrophic failure, wherein the vehicle can no longer maintain the desired path. 
A loss of communications from race control or the base station hinders the ability to bring the vehicle to a stop or shut down the engine in case of an incident on track.

\begin{table*}[h]
\begin{tabular}{|l|l|l|l|}
    \hline
    \multicolumn{1}{|c|}{\textbf{Fault Type}} & \multicolumn{1}{c|}{\textbf{Probability}} & \multicolumn{1}{c|}{\textbf{Severity}} & \multicolumn{1}{c|}{\textbf{Criticality}} \\
    \hline
    Node Health & Remote & Critical & Major \\
    Data Health & Probable & Marginal & Major \\
    Behavioral-Safety & Frequent & Critical & High \\
    \hline
\end{tabular}
\caption{\label{table:FMECA}The Probability, Severity, and Criticality for each fault type after running FMECA analysis. Probabilities range from Frequent to Remote, and Severities range from Marginal to Critical.}
\end{table*} % https://dmd.solutions/wp-content/uploads/2016/03/6-FMECA.png

\subsection{Failure Modes}
Based on the safety considerations and flow of information discussed for the control, localization, communication, and perception modules in our software stack, we conducted a failure mode analysis to categorize the failures into node health faults, data health faults, and behavioral-safety faults.
We also performed Failure Mode, Effects and Criticality Analysis (FMECA) for each type of fault, shown in table~\ref{table:FMECA}.
FMECA is a technique for analyzing a system's safety by identifying each potential point of failure and classifying its probability and its severity.
FMECA is often used to perform safety analysis in both autonomous and manned systems~\cite{fmeca_nasa,fmeca_car}, including safety analysis of code on autonomous systems~\cite{fmeca_autonomous}.
For our analysis, we defined the probability of a fault occurring as being Frequent, happening every time the vehicle is on track, Probable, happening at least once per week, Remote, unlikely to happen more than once per season, or Unlikely.
Severity can be Negligible, meaning no loss of safety on the vehicle, Marginal, meaning decreased safety but no immediate risk of a crash, or Critical, meaning a crash is likely without immediate intervention.
The Criticality is derived from the probability and severity, and can be Insignificant, Minor, Major, or High.
%There are three types of faults that can occur on the vehicle:
% \begin{enumerate}
%     \item Node Health Faults
%     \item Data Health Faults
%     \item Behavioral Health Faults
% \end{enumerate}

\noindent \textbf{1) Node Health Faults:}
A node health fault is a system-level fault that occurs when a safety-critical process (ROS2 node) crashes, stalls, or becomes unresponsive.
In our safety stack, we consider the control module as being the most critical for safe operation of the vehicle.
The Long Control Node (Section \ref{sec:long_control}) controls acceleration and braking while the Path Tracker Node (Section \ref{sec:path_tracker}) controls steering, and a failure in either is catastrophic as it directly affects the ability of the vehicle to navigate safely.
The SSC Interface Node is responsible for communication with the lower level Raptor DBW hardware, and a failure in this node makes it impossible to send control commands to the vehicle.
Node stalls in the perception module can be caused by ROS2 code taking too much time to iterate over densely packed point cloud data.
The frequency of node failure during driving is remote because typically bugs that could cause crashes are caught during simulation testing. However, because node failure means that we cannot control the vehicle it has a critical severity.
%The safety module nodes are responsible for ensuring the safety for our vehicle. If any of these nodes fail, we need to take action to keep the vehicle safe.
Consequentially, monitoring the health and \textit{``liveness''} for safety-critical nodes is a necessity.

\noindent \textbf{2) Data Health Faults:}
Data health can be defined as a combination of data quality, and data rate.
To avoid overlap with node health faults, we only consider data health faults that do not escalate to a node crash.
A data health fault can be thought of as a sensor-level fault that may occur when data reported from a vehicle sensor is inaccurate or updated data is not being reported in a timely manner.
As an example, consider a critical sensor - such as the GNSS. An inaccurate report of the position of the vehicle can induce errors in our localization estimate, and cause the control module to drift away from the desired path. Likewise, slower GNSS data may cause the vehicle to overcorrect the steering after the vehicle drifts off of its defined path. 
Data health faults also include limit faults - sending or receiving data that is outside of the expected range, such as attempting to send a steering angle to the actuator that falls outside of the actuator's physical range.
% These also can include Typeset faults - i.e. publishing an incorrect datatype message on a ROS topic could in fact cause the ROS node to crash.
These also can include faults caused by the DDS middleware, such as messages arriving late or failing to arrive at all.
Another example of a data fault for our stack is failure to receive race flag updates from race control, or failing to receive updates from the base station. These periodic \textit{pings} from MyLaps race control and the Pits base station are needed to maintain the ability to bring the vehicle to an emergency stop if necessary.
Data Health Faults are much more common than Node Health Faults. For example, the vehicle may fail to receive updates from MyLaps or the base station for multiple consecutive seconds each lap if some part of the track has poor radio coverage.
The severity is marginal compared to Node Health Faults because the code can still control the vehicle, and there are multiple ways to stop the vehicle in an emergency (e.g., if the vehicle fails to receive updates from MyLaps it can still be stopped from the base station).

\noindent \textbf{3) Behavioral-Safety Faults:}
A behavioral-safety fault is an algorithmic-level fault that occurs when the vehicle behaves in an unsafe manner due to poor programming, incorrect logic, or bad implementation of an algorithm or a routine.
Behavioral-safety faults occur when the raw data from the sensors is considered healthy, but the algorithms using the healthy data make a wrong decision.
In the context of our autonomous racing setup, we consider two critical behavioral-safety faults. 
The first is failure to obey race flags.
The second behavioral-safety fault is failure to provide adequate separation when overtaking another vehicle. Recall that after completing an overtake, the leading vehicle must leave a longitudinal separation of at least 30m before ``\textit{closing-the-door}'' on the opponent and merging back to the raceline. 
%This separation is necessary to keep both vehicles safe in the event of an unexpected situation, such as an emergency stop from one vehicle.
Both requirements are described in detail in Section~\ref{sec:setup}.
Behavioral-Safety Faults are potentially the most frequent. For example, during an overtake, the perception algorithm may have missed detections of the opponent every few seconds. In addition to being frequent, they also have a critical severity as they could cause a multi-vehicle crash with little time to react.

\subsection{Safety Archetypes and Monitoring Nodes}
Each type of safety fault needs to be handled by some node to have an assurance of vehicle safety on the track.
A team's ability and comfort level in sending their race cars to the track full-autonomously and unsupervised is directly a function of their confidence in their ability to handle these faults. 
\begin{algorithm}
\caption{Node Health Monitoring}\label{alg:code_node_health}
\DontPrintSemicolon
\SetKwFunction{HBCallback}{heartbeatCallback}
\SetKwFunction{TimerCallback}{timerCallback}

\SetKwProg{Fn}{Function}{:}{}
\Fn{\HBCallback{hearbeat\_msg}}{
    heartbeat\_value = heartbeat\_msg.data \\
    heartbeat\_time = now()
}

\SetKwProg{Fn}{Function}{:}{}
\Fn{\TimerCallback{}}{
  time\_since\_heartbeat = now() - heartbeat\_time \\
  \If{time\_since\_heartbeat $>$ safety\_threshold}
    {
    stop\_vehicle = true \\
    publish(stop\_vehicle)
    }
}
\end{algorithm}
Node health faults can be handled by monitoring that the safety-critical nodes are still operating and producing data at the expected rate.
Similar to a watchdog monitor, this can be achieved by having a node produce a heartbeat signal for the monitor, and the monitor continuously checking the heartbeat as evidence of the node still being operational.
One version of the heartbeat is in the form of a boolean flag, published at a set rate.
Algorithm \ref{alg:code_node_health} gives an example of a Node-health monitor that operates by detecting boolean flag heartbeats.
In this algorithm, the node uses a timer to compare the current time against the last time a heartbeat was received. If the difference between these two times is over a set threshold (defined as \textit{safety\_threshold}), then the node stops the vehicle.
Another variant of heartbeat detection is to have the heartbeat be an integer that increments with each beat. This allows the monitor to detect which heartbeats and how many were not received by recording the value of each heartbeat as it arrives.

% \begin{lstlisting}[caption={Pseudocode of a simple node heartbeat checking algorithm}, label=lst:code_node_health]
% function heartbeatCallback(heartbeat_msg){
%   heartbeat_value = heartbeat_msg.data
%   heartbeat_time = now()
% }
% function timerCallback(){
%   time_since_heartbeat = now() - heartbeat_time
%   if (time_since_heartbeat > safety_threshold) {
%     stop_vehicle = true
%     publish(stop_vehicle)
%   }
% }
% \end{lstlisting}

% \begin{algorithm}[H]
% \DontPrintSemicolon
% \SetKwFunction{HBCallback}{heartbeatCallback}
% \SetKwFunction{TimerCallback}{timerCallback}

% \SetKwProg{Fn}{Function}{:}{}
% \Fn{\HBCallback{hearbeat\_msg}}{
%     heartbeat\_value = heartbeat\_msg.data \\
%     heartbeat\_time = now()
% }

% \SetKwProg{Fn}{Function}{:}{}
% \Fn{\TimerCallback{}}{
%   time\_since\_heartbeat = now() - heartbeat\_time \\
%   \If{time\_since\_heartbeat $>$ safety\_threshold}
%     {
%     stop\_vehicle = true \\
%     publish(stop\_vehicle)
%     }
% }

% \caption{Boolean heartbeat checking}\label{alg:code_node_health}
% \end{algorithm}
\begin{algorithm}
\caption{Data Health Monitoring}\label{alg:code_data}
\DontPrintSemicolon
\SetKwFunction{ReceiveData}{receiveDataCallback}

\SetKwProg{Fn}{Function}{:}{}
\Fn{\ReceiveData{data\_msg}}{
  data\_value = data\_msg.data \\
  data\_accuracy = data\_msg.stddev \\
  \If{data\_value $<$ lower\_bound \textbf{or} data\_value $>$ upper\_bound}
    {
    \KwRet
    }
  \ElseIf{data\_accuracy $>$ accuracy\_threshold}
    {
    \KwRet
    }
  \Else
    {
      main\_routine(data\_value)
    }
}
\end{algorithm}
Data health faults are handled by monitoring the accuracy and frequency of the data.
The frequency of the data can be checked in a manner similar to detecting node heartbeats, by keeping track of the time since the data message was received.
The accuracy of the data can be monitored in several ways. 
If the data provides an accuracy estimate of its own (for instance the GNSS data), such as standard deviation, that value can be monitored directly.
If accuracy is not reported in the data, a routine can be implemented to track the previous values of the reported data, and detect when an unexpected change occurs.
Another alternative is to compare the data from multiple sensor sources.
The data health monitor can also implement a check on the limits of the data, and reject any data that falls outside of the expected range.
Algorithm \ref{alg:code_data} shows an example of a data health monitor based on accuracy and data limit checking.
The algorithm checks if the data (\textit{data\_value}) is outside of the defined bounds (\textit{lower\_bound} and \textit{upper\_bound}).
It also checks if the accuracy (\textit{data\_accuracy}) is above the defined threshold (\textit{accuracy\_threshold}).
Only if the data is healthy does the node use the data in its main routine.

% \begin{lstlisting}[caption={Pseudocode of a simple data checking algorithm}, label=lst:code_data]
% function receiveDataCallback(data_msg){
%   data_value = data_msg.data
%   data_accuracy  = data_msg.stddev
%   if (data_value < lower_bound || data_value > upper_bound){
%     return
%   } else if (data_accuracy > accuracy_threshold) {
%     return
%   } else {
%     main_routine(data_value)
%   }
% }
% \end{lstlisting}

Behavioral-safety faults are handled by having a monitor for the MyLaps race flag compliance, and obstacle detection and tracking.
%, and issues speed changes and path changes accordingly.
This monitor needs to set the desired velocity according to the race flag. %including knowing to set the yellow flag speed lower while on pit lane.
While doing so, it is important not to continuously send desired velocity commands, as that would prevent any other safety-node from changing the desired velocity of the vehicle.
The monitor should only set the speed when the flag changes colors, to ensure that the speed only gets set once for each flag.
A example algorithm for this flag logic is shown in Algorithm \ref{alg:code_behavioral}.
The algorithm checks the new flag against the previous flag, and only if the flag has changed to a new color does it publish a new desired velocity.
In addition to speed changes, the behavioral node should monitor obstacles to avoid a crash.
The obstacle monitoring must ensure that false positive and false negative detection does not cause the vehicle to behave erratically, as described in Section VII.

% \begin{lstlisting}[caption={Pseudocode for changing speeds in response to receiving a flag}, label=lst:code_behavioral]
% function flagCallback(flag_msg){
%   prev_flag = current_flag
%   current_flag = flag_msg
%   if (current_flag == red && prev_flag != red){
%     desired_velocity = red_flag_speed
%     publish(desired_velocity)
%   } else if (current_flag == yellow && prev_flag != yellow){
%     desired_velocity = yellow_flag_speed
%     publish(desired_velocity)
%   } else if (current_flag == green && prev_flag != green){
%     desired_velocity = green_flag_speed
%     publish(desired_velocity)
%   }
% }
% \end{lstlisting}

\begin{algorithm}
\caption{Behavioral-Safety Monitor: Race Flags}\label{alg:code_behavioral}
\DontPrintSemicolon
\SetKwFunction{FlagCallback}{flagCallback}

\SetKwProg{Fn}{Function}{:}{}
\Fn{\FlagCallback{flag\_msg}}{
  prev\_flag = current\_flag \\
  current\_flag = flag\_msg \\
  \If{current\_flag == red \textbf{and} prev\_flag != red}
    {
    desired\_velocity = red\_flag\_speed \\
    publish(desired\_velocity)
    }
  \ElseIf{current\_flag == yellow \textbf{and} prev\_flag != yellow}
    {
    desired\_velocity = yellow\_flag\_speed \\
    publish(desired\_velocity)
    }
  \ElseIf{current\_flag == green \textbf{and} prev\_flag != green}
    {
    desired\_velocity = green\_flag\_speed \\
    publish(desired\_velocity)
    }
}
\end{algorithm}

%% file: sections/safety_architecture.tex
We now present the HALO safety architecture for our software stack. 
HALO comprises of a combination of nodes working alongside the main software stack, which ensure that the race car operates safely at all times. 
The composition of HALO is derived based on the failure mode and criticality analysis, as well as the safety archetypes presented in Section \ref{sec:safety}.
While we present HALO within the context of our implementation for a high-speed AV-21 race car, several of the HALO nodes can be generalized to other autonomous driving stacks. 
The HALO safety architecture on our vehicle consists of four nodes, each of which are described next.

% The Graceful Stop Node implements a routine to safely bring the vehicle to a stop when necessary, and also implements some data health checks.
% The Node Health Monitor performs node health checks on safety-critical nodes.
% The Topic Multiplexer node performs data health checks to determine the best source when multiple data inputs are available.
% Lastly, the SSC Interface performs behavioral health checks for our software stack.

\subsection{HALO Node I: Graceful Stop}
\label{sec:graceful_stop}
The Graceful Stop Node's role is to bring the vehicle to a stop in a safe and controlled manner.
This is different from an emergency stop which immediately locks the brakes and shuts off the vehicle's engine.
An emergency stop is very risky and a considered as a last resort to stop the vehicle. Locking the brakes at high-speeds can cause the vehicle to immediately lose traction and spin out. 
%This behavior is not desired, as locking the brakes can damage the drivetrain. % TODO: Is this the reason why we want to avoid locking the brakes?
%Also, restarting the vehicle after shutting off the engine takes a long time.
Instead, the Graceful Stop Node brings the vehicle to a stop in a more gradual manner.
When the Graceful Stop Node detects that the vehicle needs to be brought to a stop, it sends a boolean data flag to the Long Control Node (Section \ref{sec:long_control} - responsible for maintaining a desired velocity). The Long Control Node, upon receiving this boolean data flag, sets the desired speed of the vehicle to zero, and thus brings the vehicle to a gradual safe stop.
%This allows the vehicle to come to a stop without taking control of the accelerator and brake away from the Long Control Node.
In addition to setting the desired velocity, the Long Control Node also sets an internal flag - as long as this internal flag is in effect, the Long Control Node will not change the desired speed from zero.
To remove this flag, the long control node must not receive a stop flag from graceful stop for a set amount time. %(5 seconds).
In our architecture, we set this reset period to 5 seconds -- long enough to ensure that the problem that caused the graceful stop is resolved.
This behavior prevents the race car from unexpectedly exiting the graceful stop mode.

\begin{figure}
     \centering
     \includegraphics[width=\columnwidth]{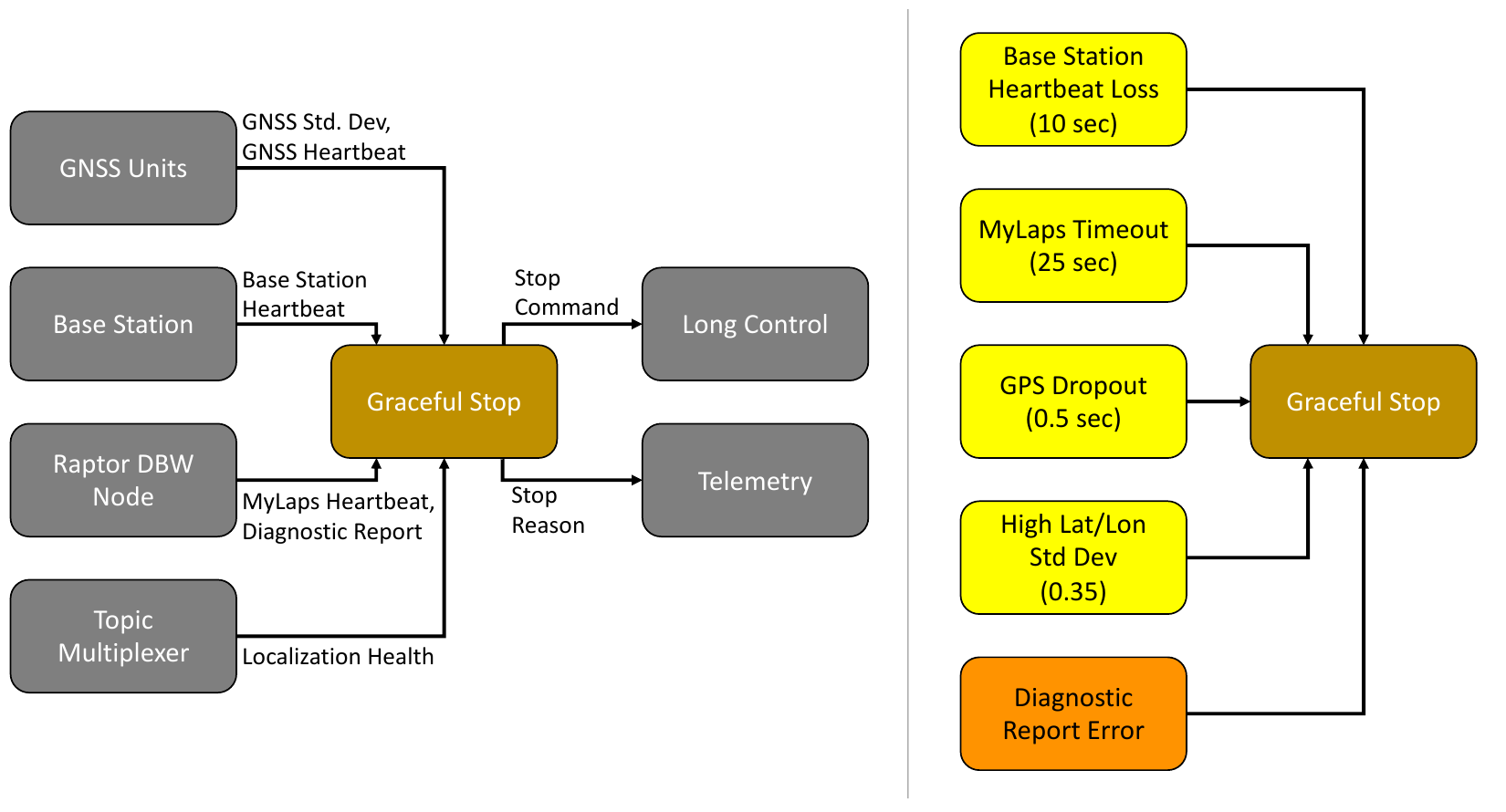}
     \caption{The Graceful Stop Node provides data health monitoring, and brings the vehicle to a gradual stop upon detecting a fault. Each data health fault has its own severity. Yellow represents recoverable faults, while orange represents non-recoverable faults.}
     \label{fig:graceful_stop}
 \end{figure}

In addition to implementing the graceful stop routine, the Graceful Stop Node also applies data health monitoring on critical data, which can be seen in Figure \ref{fig:graceful_stop}. 
If any of these checks is failed, the Graceful Stop Node will bring the vehicle to a stop.
%One such check is on the MyLaps race flag data. The Graceful Stop Node detects if it has not received a flag from race control in some time (25 seconds), and if so it brings the vehicle to a stop.
One such check is whether the MyLaps race flag data has not been received in some time. % (25 seconds).
We set this timeout to a high value, 25 seconds, to be long enough for the vehicle to be able to drive through any MyLaps dead zones.
This may occur if the MyLaps system has an outage in some sector of the track.
Similarly, the Graceful Stop Node implements a check on whether the base station heartbeat data has not been received in some time (10 seconds -- long enough to drive through any mesh radio network dead zones).
%This could be the case if the vehicle network bandwidth is flooded with traffic, so the base station cannot communicate with the vehicle.
The Graceful Stop Node also checks if the GNSS data has not arrived for some time. % (500 milliseconds).
In other domains, it may be feasible to drive for a longer time without receiving GNSS data, but because of the high-speed nature of autonomous racing, we require decimeter-level accuracy in our localization to avoid a crash.
Therefore, we chose to set this value to 500 milliseconds because at a typical racing speed of 40 mps this equates to driving 20 meters, which is the longest distance we feel safe driving using the EKF estimated localization without an updated ground truth from the GNSS units.
Additionally, the graceful stop node checks the accuracy of the GNSS data using the latitude's and longitude's standard deviation reported as part of the GNSS data. Both of these values must be below a set threshold (35 centimeters) to continue driving.
The 35 centimeter threshold was set because at higher values the vehicle is at risk of hitting a wall or being in the wrong lane during an overtake.
%A case where these values could exceed 35 centimeters is if the RTK corrections stop being applied to the GNSS data.
The final data check is on a diagnostic report from the low-level hardware. This diagnostic report indicates an issue with the low-level hardware, in which case, the Graceful Stop Node brings the vehicle to a stop.

The data faults checked by the Graceful Stop Node do not all have the same level of severity, as shown in Figure \ref{fig:graceful_stop}. 
%Most of these faults can be recovered from.
The vehicle may continue operation after a base station heartbeat loss or a MyLaps timeout by restoring communication with the base station or race control respectively.
Similarly, the vehicle may recover from a GNSS dropout by restoring communication with the satellites.
A high GNSS standard deviation can be solved by restoring communication with the RTK correction service.
The most severe fault checked by the Graceful Stop Node is the diagnostic report errors.
In the event of a diagnostic report error it is advised to abandon the run and bring the vehicle back to the pits for more in-depth diagnosis.
%, as these errors indicate a failure in the DBW Raptor System.

 % \begin{figure}
 %     \begin{subfigure}{0.49\textwidth}
 %         \centering
 %         \includegraphics[width=\columnwidth]{figs/graceful_stop.pdf}
 %         \caption{The Graceful Stop Node provides data health monitoring, and brings the vehicle to a gradual stop upon detecting a fault.}
 %         \label{fig:graceful_stop}
 %     \end{subfigure}
 %     \hfill
 %     \begin{subfigure}{0.49\textwidth}
 %         \centering
 %         \includegraphics[height=4.8cm]{figs/gstop_severity.pdf}
 %         %\includegraphics[width=\columnwidth]{figs/gstop_severity.pdf}
 %         \caption{The severity of faults occurring in the graceful stop node. Yellow represents recoverable faults, while orange represents non-recoverable faults.}
 %         \label{fig:graceful_stop_severity}
 %     \end{subfigure}
 %     \caption{An overview of the Graceful Stop Safety Node}
 % \end{figure}

%  \begin{figure}
%      \centering
%      \includegraphics[width=\columnwidth]{figs/gstop_severity.pdf}
%      \caption{The severity of faults occurring in the graceful stop node. Yellow represents recoverable faults, while orange represents non-recoverable faults.}
%      \label{fig:graceful_stop_severity}
%  \end{figure}

\subsection{HALO Node II: Node Health Monitor}
\label{sec:node_health_monitor}
\begin{figure}
     \begin{subfigure}[t]{0.49\textwidth}
         \centering
         \includegraphics[width=\columnwidth]{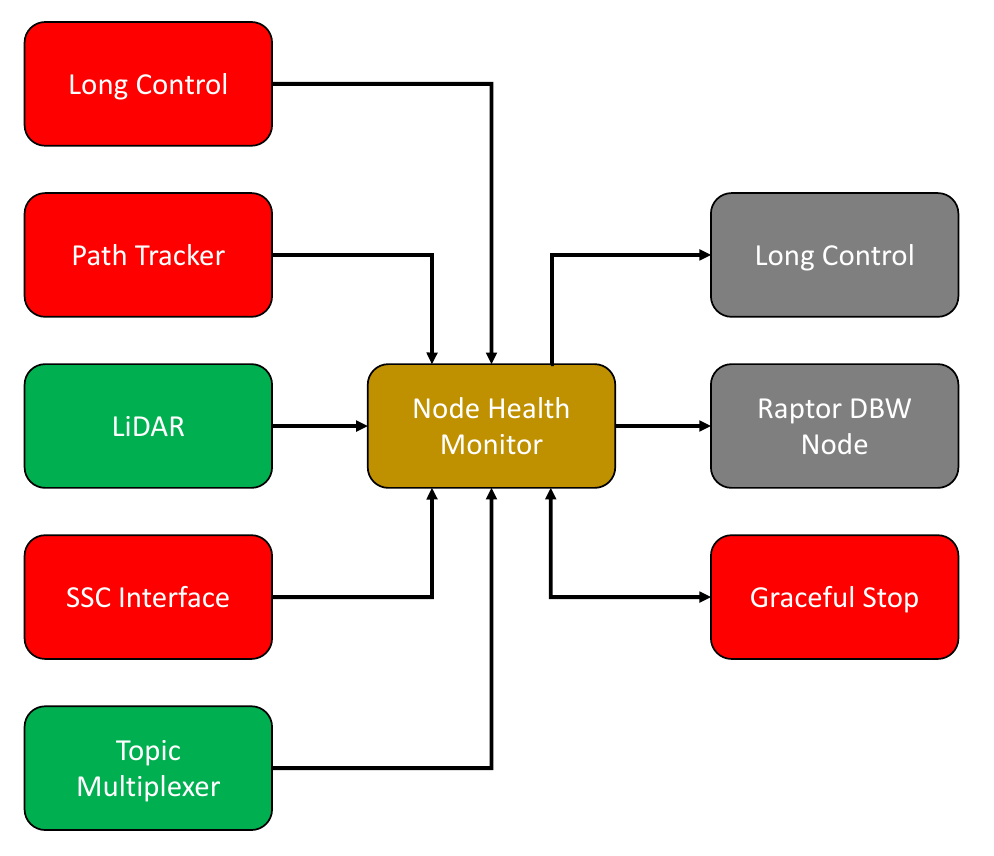}
         \caption{The severity of a node health fault occurring. Red indicates a node crash that leads to an unrecoverable emergency stop, while green represents a node crash that may be able to be ignored depending on the circumstances. Grey nodes are those used to perform an emergency stop if a graceful stop is not possible.}
         \label{fig:node_health_monitor}
     \end{subfigure}
     \hfill
     \begin{subfigure}[t]{0.49\textwidth}
         \centering
         \includegraphics[width=\columnwidth]{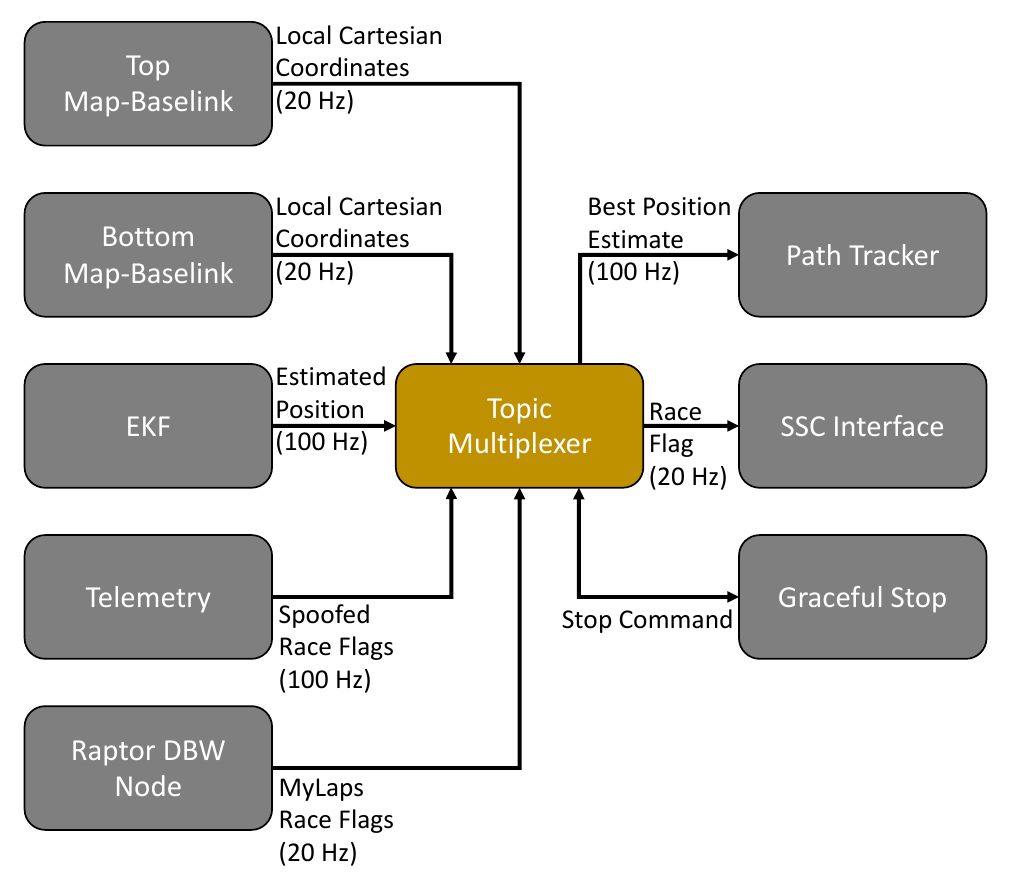}
         \caption{The Topic Multiplexer Node chooses between multiple data sources to provide the remaining nodes with the best data available. It also detects data health faults in the data it processes.}
         \label{fig:topic_multiplexer}
     \end{subfigure}
     \caption{An overview of the Node Health Monitor and Topic Multiplexer Safety Nodes.}
 \end{figure}

The Node Health Monitor node is responsible for handling node health faults in our stack.
This node monitors the safety-critical nodes in our stack, which can be seen in Figure \ref{fig:node_health_monitor}.
%, and stops the vehicle if any one of them fails.
The nodes that are considered safety-critical in our architecture are Long Control, Path Tracker, SSC Interface, LiDAR, Topic Multiplexer, and Graceful Stop.
Long Control, Path Tracker, and SSC Interface, are considered safety-critical because the vehicle cannot be controlled if they fail. LiDAR is considered safety-critical because the vehicle cannot reliable detect obstacles without the LiDAR. Topic Multiplexer, and Graceful Stop are considered safety critical because a failure in either of these nodes can cause other failures to escalate.
Depending on the particular node that failed, the node health monitor is capable of stopping the vehicle in different ways.
The preferred way of stopping the vehicle is to send a stop request to the Graceful Stop Node.
%, to use the graceful stop routine to bring the vehicle to a stop.
For instance, this is the procedure in case the Path Tracker Node dies.
However, the graceful stop routine itself relies on several safety-critical nodes being operational in order to bring the vehicle to a stop. If any of the nodes that the graceful stop routine relies on has died, the Node Health Monitor must stop the vehicle in another way.
For example, if the Graceful Stop Node has died, the Node Health Monitor will mimic the Graceful Stop Node's behavior by publishing a desired velocity of zero to the Long Control Node, allowing the Long Control Node to bring the vehicle to a stop.
In the event that the Long Control Node has failed, then it will be unable to control the brakes to stop the vehicle. In this case, the Node Health Monitor will instead publish a moderate brake command of its own to the SSC Interface to bring the vehicle to a stop.
If the SSC Interface has also died, then the Node Health Monitor can no longer communicate control commands to the low level actuators. In this case there is no way to stop the vehicle in a graceful manner, and the only recourse is to stop the vehicle by shutting off the engine (Purple race flag, or through the base station joystick).
Lastly, the Node Health Monitor must also be checked since it is a safety-critical node. To implement this check, we have the Node Health Monitor produce its own heartbeat, and have the Graceful Stop node stop the vehicle if the heartbeat is not received.

In addition to stopping the vehicle in response to a node failure, the Node Health Monitor can provide other functionality as a response.
For example, if the Topic Multiplexer (Section \ref{sec:topic_multiplexer}) dies, the Node Health Monitor will assume the duties of the Topic Multiplexer by publishing the odometry data coming from the GNSS units.
%To avoid the error that caused the topic multiplexer to crash, the node health monitor does not duplicate the multiplexing functionality of the topic multiplexer.
%Instead, the Node Health Monitor always reports the data coming from the top GNSS as localization data for the control module.
%During this time, the Graceful Stop Node will continue to monitor the accuracy of the top GNSS data, and will bring the vehicle to a stop if it exceeds its defined threshold.
Another case of a node crash that does not stop the vehicle is the LiDAR Node. In the event that the LiDAR data stops getting reported, the Node Health Monitor informs the operator at the base station that this data was not received, allowing the human to determine whether to allow the vehicle to continue driving.
If the vehicle is the only vehicle on the track, it may be safe to continue driving. If there is another vehicle present, however, a stop may be necessary to keep all vehicles safe.

%  \begin{figure}
%      \centering
%      \includegraphics[width=\columnwidth]{figs/node_health_monitor.pdf}
%      \caption{The Node Health Monitor provides monitors node heartbeat data, and stops the vehicle if it detects a fault.}
%      \label{fig:node_health_monitor}
%  \end{figure}

%The severity of a node health fault depends on which node has crashed, as shown in Figure \ref{fig:node_health_severity}.
If the Long Control, Path Tracker, SSC Interface, or Graceful Stop Nodes crash, the vehicle can no longer continue to run, and must be manually recovered and taken back to the pits.
A crash of the topic multiplexer node is less severe, because the node health monitor will take over its role of providing localization data to the control module. 
%However, since the localization data will be at the 20 Hz lower rate, it is inadvisable to drive at high speeds.
%The severity of the LiDAR Node crashing depends on the circumstances of the track, and it is up to the human operator at the base station to decide whether to bring the vehicle to a stop.

%  \begin{figure}
%      \centering
%      \includegraphics[width=\columnwidth]{figs/node_health_severity.pdf}
%      \caption{The severity of a node health fault occurring. Red indicates a node crash that leads to an unrecoverable emergency stop, while green represents a node crash that may be able to be ignored depending on the circumstances.}
%      \label{fig:node_health_severity}
%  \end{figure}

\subsection{HALO Node III: Topic Multiplexer}
\label{sec:topic_multiplexer}
The Topic Multiplexer node is essentially a data health monitor that chooses between different data sources and provides whichever is most accurate. 
The data monitored by this node can be seen in Figure \ref{fig:topic_multiplexer}.
The node continuously determines the most accurate source of localization between the two GNSS units and the EKF estimation.
Because the EKF position is reported at a higher rate of 100 Hz, the topic multiplexer defaults to EKF as the mian source of localization.
However, the EKF could become inaccurate even while the GNSS remains accurate, such as in the case where the vehicle makes an unexpected sharp turn.
Therefore, if the EKF covariance is above a threshold of 0.1225, then the topic multiplexer falls back on the slower 20 Hz GNSS as the source of localization.
The covariance threshold of 0.1225 is the square root of the lateral standard deviation of 35 centimeters, making the safety margin the same for both the raw GNSS and the EKF estimation.
To determine which GNSS is more accurate, the node monitors the covariance, which is derived from the standard deviation reported with the latitude and longitude.
In order to switch back from the lower rate GNSS to the faster EKF estimation, the node requires some number of EKF messages in a row which all have a covariance below the defined threshold. 
%In our architecture, 5 messages are required to return to using EKF data.
%In addition to detecting bad data through the covariance, the node also detects when data has stopped coming from any of these sources.
If any source stops reporting data, the topic multiplexer will disregard that source when determining the most accurate data.
If all sources stop reporting data at the same time, the topic multiplexer will send a flag to the graceful stop node to bring the vehicle to a stop. This could be the case if both GNSS units lose connection with the satellites.

% In addition to choosing between localization sources, the topic multiplexer also has the ability to switch between sources of race flags.
% This switch occurs based on setting a parameter of the topic multiplexer node, rather than being automatically detected. This is done intentionally to inform the base station when the vehicle is no longer receiving flags from race control.
% The base station operator may choose to switch from receiving flags from race control to receiving spoofed flags from the base station.
% This removes a point of failure by removing race control's ability to stop the car in an emergency, but it allows the vehicle to continue driving in the event that the race control infrastructure is unable to send flags to part or all of the track.

%  \begin{figure}
%      \centering
%      \includegraphics[width=\columnwidth]{figs/topic_multiplexer.pdf}
%      \caption{The Topic Multiplexer Node chooses between multiple data sources to provide the remaining nodes with the best data available. It also detects data health faults in the data it processes.}
%      \label{fig:topic_multiplexer}
%  \end{figure}

\subsection{HALO Node IV: Behavioral-Safety Monitor}
\label{sec:behavioral_safety}
The behavioral-safety monitor in our stack is implemented in the SSC Interface Node.
This node is responsible for changing the desired velocity of the vehicle in response to flags received from race control.
It also determines which reference path is provide to the Path Tracker Node (Section \ref{sec:path_tracker}).
%should follow.
%These decisions are the considered the behavioral health of the vehicle.

\begin{figure*}
     \centering
     \includegraphics[width=0.9\linewidth]{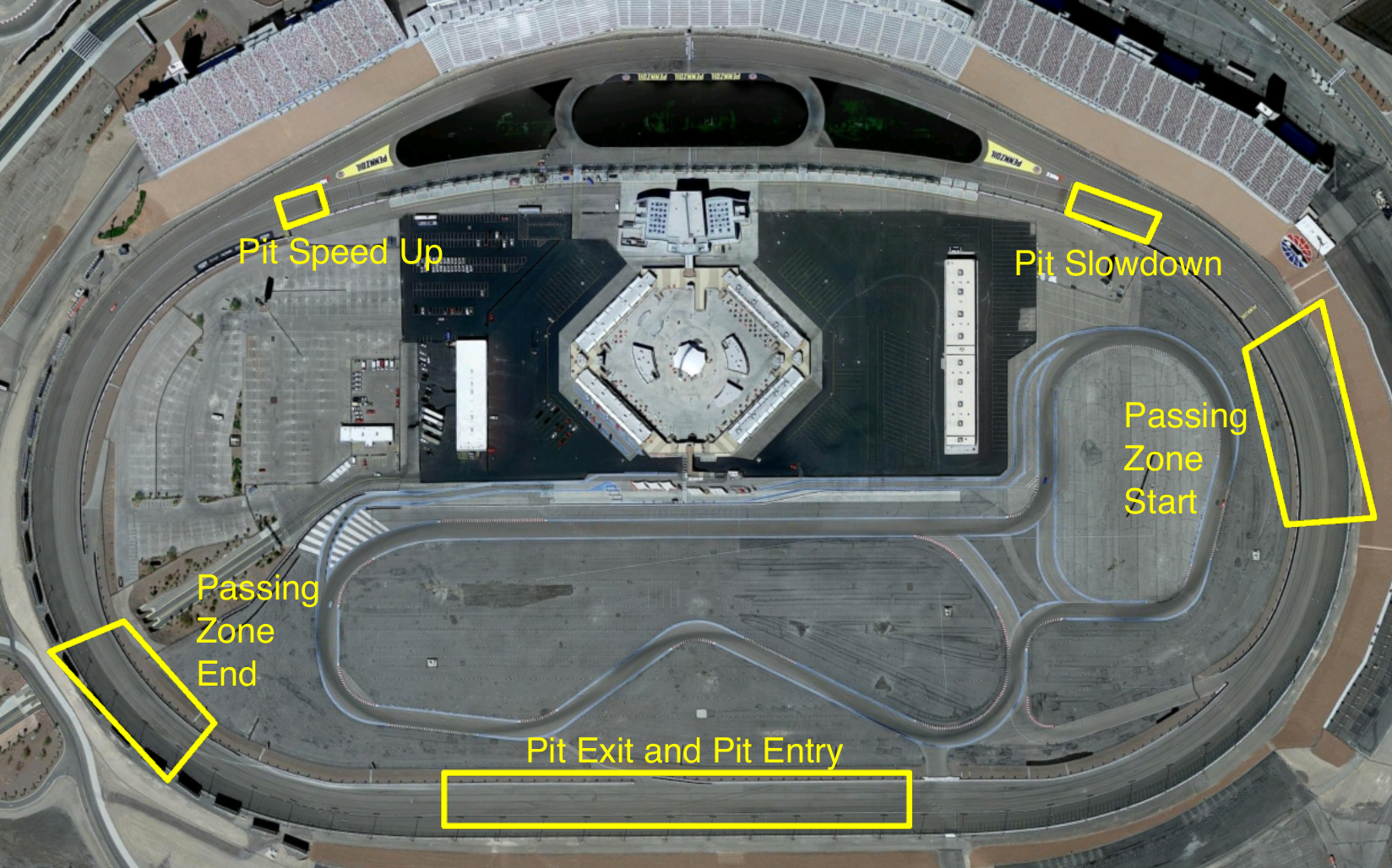}
     \caption{The transition polygons defined on the Las Vegas Motor Speedway Track. Using these polygons, the vehicle can detect whether it is in the pits or on the track (Pit Exit and Pit Entry), beside the pit boxes (Pit Slowdown and Speed Up), or if it is in the passing zone (Passing Zone Start and End).}
     \label{fig:lvms_polygons}
 \end{figure*}

\noindent \textbf{Polygon-Based Transitions for Race Flags}
The SSC Interface Node determines the desired velocity of the vehicle based the current race flag and the vehicle's location.
For example, the vehicle should be slower while driving on pit lane than while driving on the track.
The vehicle's position in this context is a coarse-grained determination of which part of the race track the vehicle is in, rather than an x,y coordinate value.
To achieve this coarse-grained localization, polygons that mark the entrance to a new section of the track have been defined in the local coordinate frame.
Figure \ref{fig:lvms_polygons} Shows the polygons that were used for the Las Vegas Motor Speedway track.
The polygon at the bottom center of the image was used to identify when the vehicle had left the pits and entered onto the track, as well as when the vehicle was leaving the track to return to the pits.
Additional polygons are used to define the start and end of the passing zone, and 
%Polygons are also used to define
the start and end of the pit lane.
%, as the vehicle must drive slower when passing the pit boxes.
%As the vehicle drives, its x,y position is checked to determine if it has entered a one of these polygons, marking a transition to the new track section.
This polygon-based transition method allows the node to know whether the vehicle is in the pit lane, in an overtaking zone, or elsewhere the track.
Using this knowledge, whenever the node receives a new flag it uses a lookup table to determine how fast the vehicle should be traveling for the part of the track.
%that it is in.
The SSC Interface Node is also responsible for changing the reference path for the path tracker to follow.

%\noindent \textbf{Path Switching}
%, between the paths defined in Section \ref{sec:control}.
%There are four instances in which the path changes. The first is when leaving pit lane and merging on the race track, the node switches from the ``pits'' path to the ``raceline'' path. Similarly, when under a black flag the node switches to the ``pits'' path when approaching the pit entry.

\noindent \textbf{Closing-the-Door Maneuver}
\label{sec:close_the_door}
%The other two cases for a path change are when attempted a vehicle overtake.
If the ego vehicle receives the signal from race control that it can overtake the opponent (Waving Green flag), it changes from the ``raceline'' path on the inside of the track to the ``overtake'' path on the outside of the track.
The vehicle will then remain on the ``overtake'' path until it detects that it has passed the opponent to the point where it has maintained the required longitudinal separation of 30m to the opponent (as defined in Section \ref{sec:setup}). 
At which time it may merge to the ``raceline'' path.
This maneuver to return to the ``raceline'' path is referred to as ``Closing-the-Door.''
% What makes this maneuver challenging is that it is inherently unsafe--the ego vehicle must take a path that puts it directly in front of the opponent vehicle.
What makes this maneuver challenging is that it requires some risk - the ego vehicle must complete this maneuver within a certain time frame in order to win the race.
In order to balance this risk and make the maneuver as safe as possible, we only close the door if we detect that we are sufficiently far in front of the other vehicle (in this case 30m).
Due to uncertainty associated with the obstacle distance reported by the perception module, it is not prudent to immediately switch paths on the first indication of the 30m distance measurement. This could be a false positive reading.
Instead, the SSC interface implements a sliding window over the reported distances, and only closes the door if at least $n$ out of the latest $k$ distance measurements are reported to be outside of the safety distance requirement.
%In our architecture, we required the vehicle to wait for 8 out of the latest 10 distance measurements to be higher than 30 meters.
%This sliding window approach prevents a single bad LiDAR scan from causing the vehicle to prematurely switch paths, which could cause a loss of separation with the opponent vehicle.

%% file: sections/results.tex
During the Indy Autonomous Challenge, we recorded ROS Bags containing all of the ROS topics and sensor data from each run at the IMS and LVMS racetracks.
Due to the limited amount of time available to us at each track, we were unable to rigorously test each module in many different experimental setups.
Nevertheless, we have collected data which highlights the effectiveness of the HALO architecture to successfully mitigate the different failure modes. 
Specifically, we present three results, each corresponding to a different failure mode described in Section \ref{sec:safety}.

\subsection{Mitigating Data Health Faults}
Figure~\ref{fig:ekf_drift}, shows an example of a data health fault.
Each block represents a node or module with a vertical arrow representing time.
Each arrow going across between these node timelines denotes a message sent between the nodes.
Here, the GNSS is reporting the vehicle's position to the EKF node and the Topic Multiplexer at 20 Hz.
The EKF Node fuses GNSS data with IMU data and velocity data to produce localization estimates at 100 Hz, and reports them to the HALO Node III: Topic Multiplexer (Section \ref{sec:topic_multiplexer}).
The Topic Multiplexer determines the best (accurate and fast) source of localization and reports it to the Control Module. Here it chooses the EKF source because the reported data covariance is below the defined accuracy threshold.
This is nominal operation of the vehicle as it drives around the track.
At some point, the covariance of the EKF localization estimates begin to rise. This may occur due to errors in the velocity estimates or IMU data.
%vehicle begins to turn too sharply for the speed it is moving at. Because of this maneuver, the vehicle begins to slide.
%As discussed in Section \ref{sec:local}, The EKF module uses the wheel speed to estimate the vehicle's speed. When the vehicle moves in a way that is inconsistent with the speed reported by the wheels, the EKF algorithm becomes inaccurate.
In Figure \ref{fig:ekf_drift}, this is shown by the red arrows emanating from the EKF, which have a high covariance value (0.12385).
During this time, even though the EKF becomes inaccurate, the GNSS data continues to be accurate, as shown by the low covariance value (0.000069).
The topic multiplexer detects the deteriorating EKF accuracy, and switches to the lower rate (but higher accuracy) GNSS data.
%, because the EKF estimate can no longer be trusted.
As seen in Figure \ref{fig:ekf_drift}, after the EKF's covariance rises, the rate of data going to the control module slows to the 20 Hz rate of the GNSS.

This example shows successful mitigation of a data health fault by the HALO Node III: Topic Multiplexer.
Without the intervention of the Topic Multiplexer, the localization covariance would have continued to increase resulting in a risk of a vehicle crash.
%vehicle would have to stop when the accuracy of the EKF data rises above the defined threshold.
Instead, the HALO Topic Multiplexer switches to the more accurate GPS data, allowing the vehicle to continue to drive safely, albeit slightly more slowly due to the lower GNSS rate of 20 Hz.

\subsection{Mitigating Node Health Faults}

 \begin{figure}
     \begin{subfigure}[][9.5cm][t]{0.49\textwidth}
         \centering
         \includegraphics[width=\columnwidth]{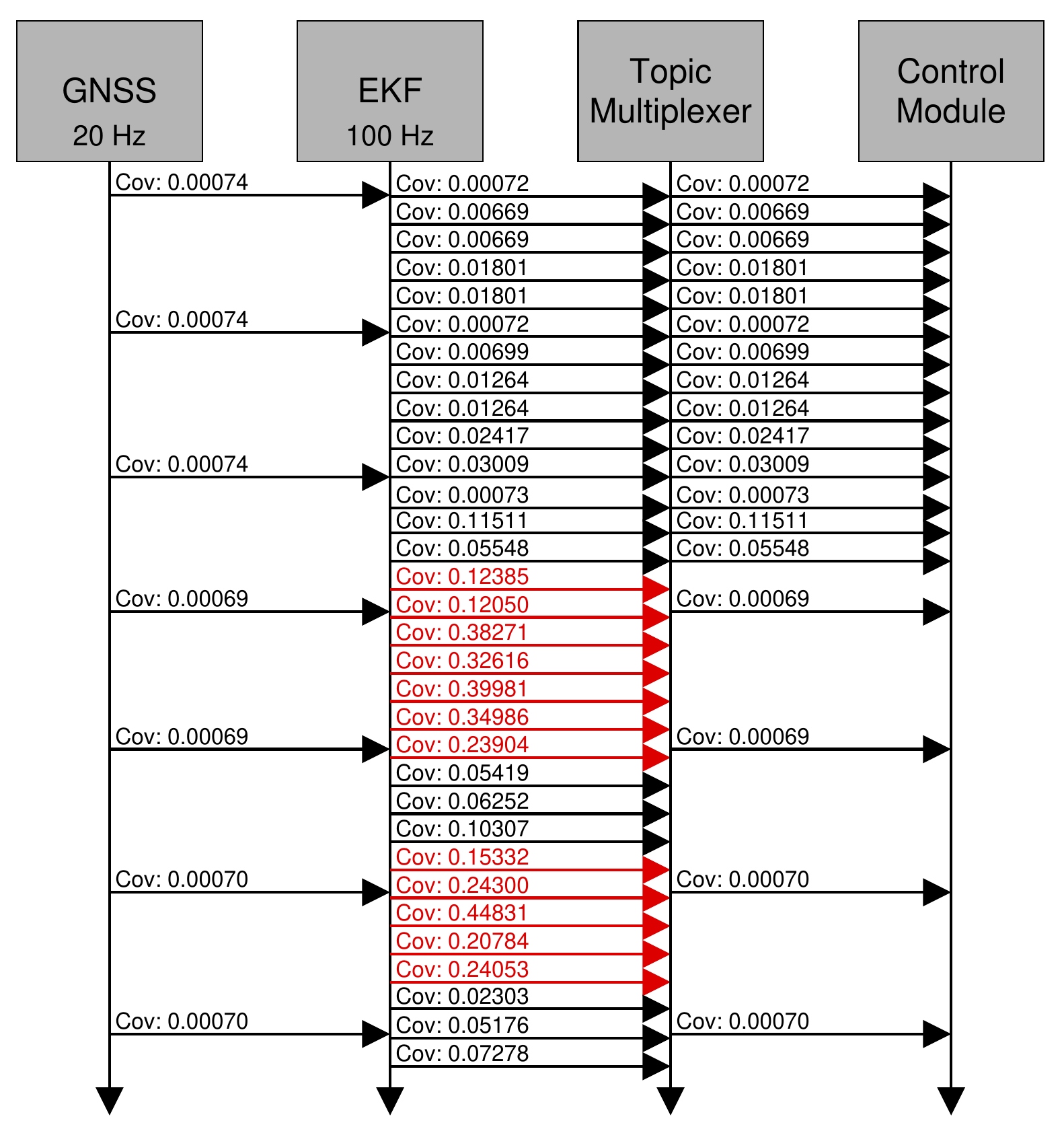}
         \caption{Data showing a data health fault. The EKF localization estimate begins to drift, but the GNSS data remains accurate. In this case, the HALO Topic Multiplexer switches from the high frequency EKF to the lower frequency, but higher accuracy, GNSS data.}
         \label{fig:ekf_drift}
     \end{subfigure}
     \hfill
     \begin{subfigure}[][9.5cm][t]{0.49\textwidth}
         \centering
         \includegraphics[width=\columnwidth]{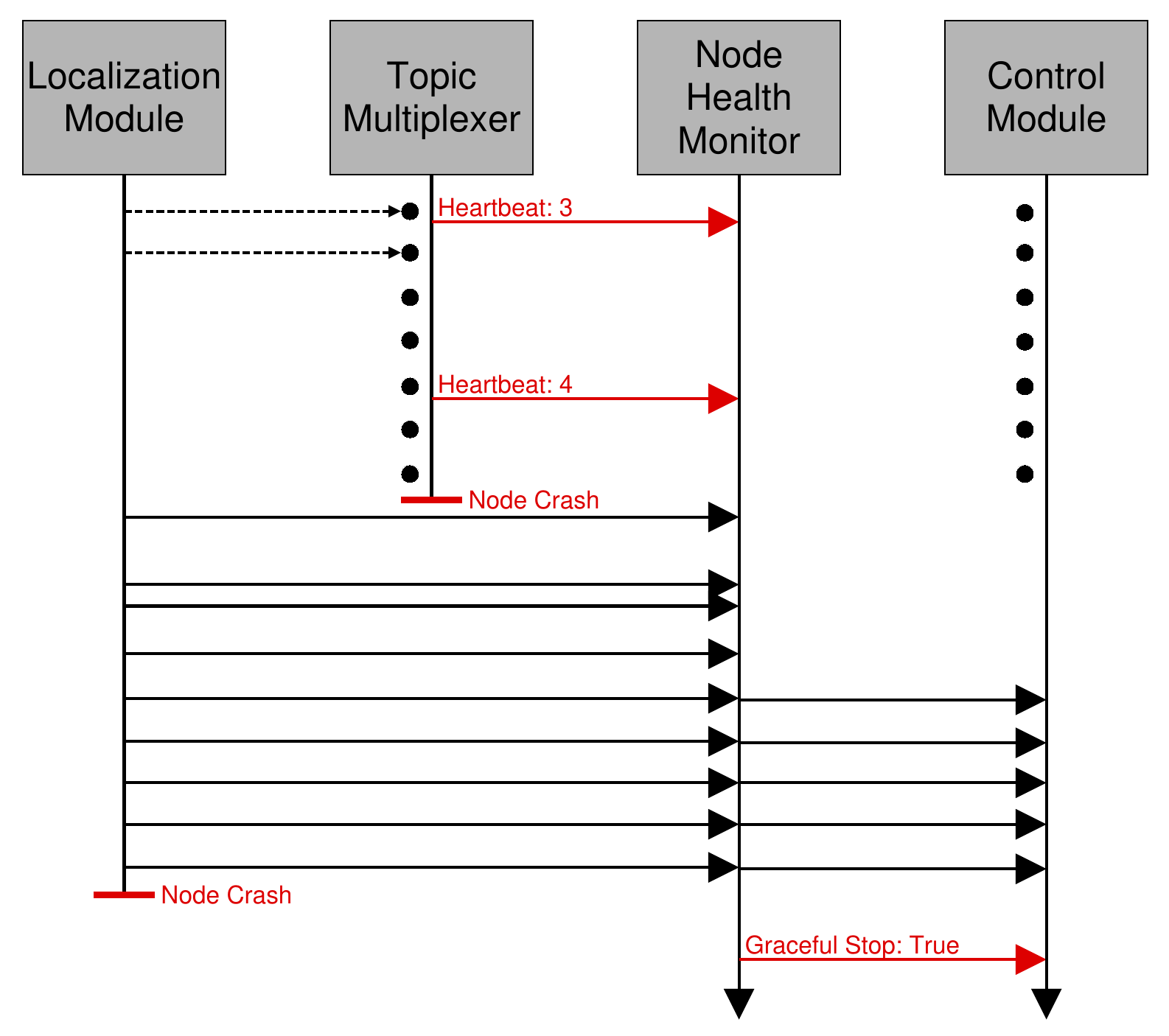}
         \caption{Data showing a node health fault. The Topic Multiplexer node crashes, so the HALO Node Health Monitor instead begins providing localization data to the control module. Then the localization node crashes, so the Node Health Monitor tells the vehicle to perform a graceful stop.}
         \label{fig:node_crash}
     \end{subfigure}
     \caption{Experimental results of two different types of faults.}
 \end{figure}
 
Figure \ref{fig:node_crash} shows an example of a node health fault, correctly mitigated by the HALO architecture.
%Again, each block represents a node or module with a vertical arrow that represents the time elapsed, and each arrow going between these node timelines represents a message being sent from a node.
In this example, the localization module is periodically reporting its position to the topic multiplexer, and the topic multiplexer is sending that data to the control module.
This data flow is shown by the black dots in Figure \ref{fig:node_crash}.
During this process, the topic multiplexer is publishing heartbeats in the form of a rolling counter to the HALO Node II: Node Health Monitor (Section \ref{sec:node_health_monitor}).
At some point the Topic Multiplexer ROS node experiences a fatal error and crashes, shown by the red bar at the end of the Topic Multiplexer's timeline.
After receiving the last heartbeat from the Topic Multiplexer, the Node Health Monitor waits to see if the node is dead or if the heartbeat message was merely delayed.
%due to insufficient compute resources.
After waiting a defined time threshold (500 milliseconds for the Topic Multiplexer), and still not receiving a heartbeat, the Node Health Monitor begins its safety protocol.
Since the Topic Multiplexer was the node which has crashed, the Node Health Monitor takes over reporting localization data to the control module.
In Figure \ref{fig:node_crash} this is shown by the arrows coming from node health monitor to the control module.
Following this switch, the node providing localization also has a fatal error and crashes.
Due to a complete lack of localization data, the Node Health Monitor tells Long Control to bring the vehicle to a stop.
In Figure \ref{fig:node_crash} this is shown by the red arrow indicating the graceful stop flag being set to \textit{true}.
This example shows the mitigation of a node health fault by the HALO Node Health Monitor.
When the Topic Multiplexer dies, the Node Health Monitor is able to keep the vehicle operational by itself providing the localization data.
Only when the localization node crashes and there is no localization data to provide to the control module does the vehicle enter into a graceful stop.

\subsection{Mitigating Behavioral-Safety Faults}

%  \begin{figure}
%      \centering
%      \includegraphics[width=\columnwidth]{figs/node_crash_short.pdf}
%      \caption{Data showing a node health fault. The Topic Multiplexer node crashes, so the HALO Node Health Monitor instead begins providing localization data to the control module. Then the localization node crashes, so the Node Health Monitor tells the vehicle to perform a graceful stop.}
%      \label{fig:node_crash}
%  \end{figure}

 \begin{figure}
     \centering
     \includegraphics[width=\columnwidth]{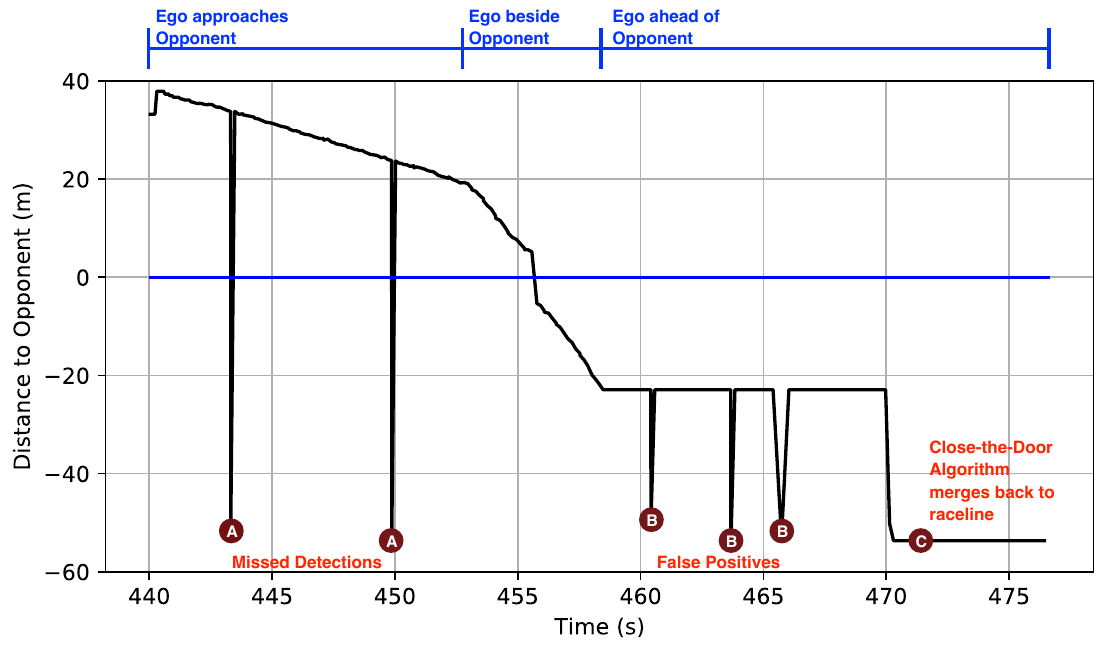}
     \caption{Data showing a HALO behavioral-safety fault. The ego vehicle approaches an opponent for an overtake. The perception stack misses several detections \circled{A} of the opponent vehicle while the opponent is ahead of the vehicle, and has several false positives \circled{B} that the opponent is safely behind the ego. Without a behavioral monitor, these missed detections would cause the vehicle to merge to the inner race line too early.}
     \label{fig:close_the_door}
 \end{figure}

%The final result we will highlight is an example of a behavioral health fault.
In this example, the Ego vehicle is overtaking an opponent. The opponent vehicle in this experiment is traveling at 60 mph, while the Ego is traveling at 80 mph.
As discussed in Section \ref{sec:setup}, the Ego needs to maintain a longitudinal separation of 30m in front of the opponent before closing-the-door. 
%, and is looking for the vehicle to be at least 30 meters behind the ego before merging back onto the inner line.
Figure \ref{fig:close_the_door} shows distance of the ego from the opponent as reported by the perception module.
%In this Figure, the distance reported is the distance from the baselink of the ego to the center of the bounding box as described in Section VII.
Positive numbers indicate that the ego is behind the opponent, while negative numbers indicate that the ego is in front of the opponent.
During the beginning of the overtake maneuver, the perception stack erroneously reports observations of the opponent far behind the ego (\circled{A} in Figure~\ref{fig:close_the_door}).
Without the HALO Behavioral-Safety Monitoring, the ego vehicle would attempt to move to the inner line as soon as the first missed detection occurs, and would merge behind the opponent's vehicle.
Since the ego is traveling faster than the opponent, merging behind the opponent could result in an increased risk of crashing into the opponent.
%With the inclusion of the Close-the-Door algorithm, the vehicle requires 8 out of the 10 last reported obstacle distances to be 30 meters behind the ego.
%Because the missed detections are sparse, the Close-the-Door algorithm detects the opponent vehicle as remaining in front of the ego, and does not attempt to merge onto the inner line.
Similarly, after the ego has pulled ahead of the opponent, there are several false positives (\circled{B} in Figure~\ref{fig:close_the_door}), where the obstacle is reported as being far behind the ego. Without the Close-the-Door algorithm described in Section~\ref{sec:close_the_door}, the ego vehicle would attempt to close-the-door on the opponent too early.
%, and would risk hitting the opponent's vehicle from the side.
Not until 470 seconds is the opponent clearly behind the ego (\circled{C} in Figure~\ref{fig:close_the_door}), and the perception stack reports the opponent's distance as consistently around -53m.
At this time, the HALO Behavioral-Safety Monitor can confidently initiate the merging maneuver.
There are other ways to ensure that the opponent is safely behind the ego besides using sliding window - for example, fusing multiple detections and using an EKF as described in Section~\ref{sec:perception} would also increase confidence in the position of the opponent's vehicle.

%% file: sections/discussion.tex
In this section, we discuss the design assumptions, limitations, and generalizability, and the definition and categorization of fault modes for the HALO architecture.
Additionally, we discuss some potential limitations of the HALO architecture and identify areas for future improvement.

In the current architecture HALO is best described as a combination of several runtime monitors - each responsible for detecting and mitigating a certain kind of fault. 
The different faults can arise in various sub-systems and components of the autonomous race vehicle. 
Note that, as mentioned earlier, the various components of the stack such as perception, planning, and control modules, and their associated algorithms can (and usually do) also have safety thresholds and other safety criteria built in, that may resemble a monitor, but the HALO architecture and the health nodes ensure that these components are functioning as desired (or required) and that any faults can be gracefully handled. 
In many fault tolerant operating system architectures, there is usually a runtime monitor, a supervisor, and safety executive layer to handle the fault. 
In the current form, the HALO health monitoring nodes are combining all three responsibilities into the health node itself. 
Therefore, the HALO health monitoring nodes continually monitor the vehicle's subsystems for faults, and decide whether to mitigate the fault (supervisor role) or take the necessary action to ensure the safety of the vehicle (safety executive role).
One limitation of the HALO safety architecture is that it assumes that all monitors are functioning correctly. If a monitor fails, it may not detect a fault, which can lead to unsafe operation. This limitation can be addressed by using techniques such as Byzantine fault tolerance~\cite{driscoll2003byzantine}, which allows the system to tolerate up to a certain number of faulty monitors.

% While the HALO safety architecture presented in this paper is specific to the autonomous racing vehicle and our racing stack, it could be generalized to other cyber-physical systems and autonomous vehicle pipelines. The architecture can serve as a starting point for developing similar fault management architectures for other systems, with modifications and customization based on the specific requirements of the system. The fault and monitor primitives are not unique to racing and can be applied and used in other autonomous CPS.
% One example of the HALO framework that could generalize to other CPS systems is the data health monitor. In general, CPS systems use multiple redundant sources for critical information. The Topic Multiplexer node described in Section~\ref{sec:topic_multiplexer} uses information from the provided data stream to automatically choose the best source of data, including choosing between processed data and raw sensor measurements if necessary. Our Topic Multiplexer node is used on the race car to switch between sources of localization data, which would be directly applicable to CPS systems such as autonomous drones that similarly require precise localization.

The HALO safety architecture presented in this paper is specific to the autonomous racing vehicle and our racing stack, and some of the concepts would not translate to other cyber-physical systems.
For example, the behavioral-safety features are tailored to the race rules and would need adaptation to be relevant for passenger cars.
However, other safety features could be generalized to other CPS and autonomous vehicle pipelines.
All vehicles which contain perception, control, localization, planning, and communication modules are vulnerable to data health and node health faults, and the procedures described in this paper could be applied to increase their safety with minimal tweaks.
The fault and monitor primitives are not unique to racing and can be applied and used in other autonomous CPS.
One example of the HALO framework that could generalize to other CPS systems is the data health monitor. In general, CPS systems use multiple redundant sources for critical information. The Topic Multiplexer node described in Section~\ref{sec:topic_multiplexer} uses information from the provided data stream to automatically choose the best source of data, including choosing between processed data and raw sensor measurements if necessary. Our Topic Multiplexer node is used on the race car to switch between sources of localization data, which would be directly applicable to CPS systems such as autonomous drones that similarly require precise localization.

% Although our framework was developed for a race car, most of the concepts apply equally to other autonomous vehicles which also contain perception, control, localization, planning, and communication modules.  
% All vehicles are vulnerable to data health and node health faults, and the procedures described in this paper could be applied to increase their safety with minimal tweaks.
% For example, because coming to a stop keeps the race car safe but would be very dangerous in a road vehicle, the graceful stop maneuver would be replaced with another behavior.
% The behavioral-safety faults are also specific to the race car, but passenger autonomous vehicles will also encounter a wide variety of behavioral-safety faults that could be mitigated using the HALO safety architecture.

The HALO safety architecture addresses three different types of faults: node health, data health, and behavioral-safety faults. Node health faults refer to failures in the subsystems of the vehicle, such as the perception, planning, control, and communication modules. Data health faults refer to issues with the data being received from sensors, such as incorrect sensor readings or sensor failures. Behavioral-safety faults refer to issues with the autonomous racing vehicle's behavior, such as failing to follow the prescribed path or violating safety distances with other vehicles. There may be some overlap between these fault modes, and a failure at a lower level (sensor failure) can propagate to a higher level (safety-behavior).
For example, a data health fault may cause a perception failure, which can result in a behavioral-safety fault. 
To address this issue, the HALO safety architecture has been designed to monitor and mitigate faults at multiple levels, from the node level to the system level, to ensure that any fault in the system is detected and mitigated before it leads to unsafe operation.

%% file: sections/conclusion.tex
%For an autonomous vehicle to be safe on a race track, it needs to be able to detect and mitigate a variety of safety faults.
This paper describes HALO, a fault-tolerant architecture for an autonomous racing software stack. 
This is among the first papers to present such a deep dive into the various faults that were experienced on a real fully autonomous race car. 
We first conduct a failure mode, effects, and criticality analysis of the control, localization, communication, and perception modules of the stack. Based on this, we present three different failure modes: data health, node health, and behavioral-safety faults.
We then implemented HALO based on our software used on a full-scale fully-autonomous AV-21 race car in the Indy Autonomous Challenge. 
Our HALO architecture comprises of a composition of nodes that are designed to handle these faults and minimize the risk of a vehicle crash, even in the presence of poor data, node crashes, and faulty perception. 
This safety comes at the cost of potentially wasted driving time, because the safety margins are designed to stop the vehicle, even if it might be safe to keep driving.
For example, the race car is safe to continue operating without MyLaps race flags as long as it is the only vehicle on track. However, driving without MyLaps means one less way to stop the vehicle in an emergency, and so we have decided to bring the vehicle to a stop until connection with MyLaps is reestablished.
%Each module has its own set of safety considerations for keeping the vehicle safe.
%These safety considerations are handled by the safety module, which aims to identify data health faults, node health faults, and behavioral health faults, and mitigate them such that the vehicle retains as much functionality as possible.
%We demonstrate the effectiveness of HALO on data gathered from the vehicle during the real-world race runs. 

% Although our framework was developed for a race car, most of the concepts apply equally to other autonomous vehicles which also contain perception, control, localization, planning, and communication modules.  
% All vehicles are vulnerable to data health and node health faults, and the procedures described in this paper could be applied to increase their safety with minimal tweaks.
% For example, because coming to a stop keeps the race car safe but would be very dangerous in a road vehicle, the graceful stop maneuver would be replaced with another behavior.
% The behavioral-safety faults are also specific to the race car, but passenger autonomous vehicles will also encounter a wide variety of behavioral-safety faults that could be mitigated using the HALO safety architecture.

While there are limitations and challenges associated with the HALO safety architecture, these can be addressed in future research, and the architecture can be extended to cover more components and scenarios. Overall, the HALO safety architecture represents an important step forward in the development of safe and reliable autonomous racing vehicles, and it has the potential to serve as a starting point for the development of safety architectures for other autonomous cyber-physical systems.
Thus, our ongoing and future work involves extending the architecture to include additional failure modes, and to integrate HALO with existing open-source self-driving stacks outside of autonomous racing.

%While driving the vehicle on-track, ROS bag data was collected, including instances where the safety module correctly identified a fault on the vehicle and mitigated the threat.

%Our ongoing and future work focuses on extending the behavioral monitor to cover a wider variety of algorithm-specific behavioral faults.
%Additionally, we aim to better decouple the safety aspects of the vehicle from the operations of the modules - for example, decoupling the base station joystick override from the operation of the control module.

%% file: main.bbl
\begin{thebibliography}{10}
\providecommand{\url}[1]{#1}
\csname url@samestyle\endcsname
\providecommand{\newblock}{\relax}
\providecommand{\bibinfo}[2]{#2}
\providecommand{\BIBentrySTDinterwordspacing}{\spaceskip=0pt\relax}
\providecommand{\BIBentryALTinterwordstretchfactor}{4}
\providecommand{\BIBentryALTinterwordspacing}{\spaceskip=\fontdimen2\font plus
\BIBentryALTinterwordstretchfactor\fontdimen3\font minus \fontdimen4\font\relax}
\providecommand{\BIBforeignlanguage}[2]{{%
\expandafter\ifx\csname l@#1\endcsname\relax
\typeout{** WARNING: IEEEtran.bst: No hyphenation pattern has been}%
\typeout{** loaded for the language `#1'. Using the pattern for}%
\typeout{** the default language instead.}%
\else
\language=\csname l@#1\endcsname
\fi
#2}}
\providecommand{\BIBdecl}{\relax}
\BIBdecl

\bibitem{aaa_survey}
``Automated vehicle survey - phase iv fact sheet,'' \emph{American Automobile Association}, Mar 2019.

\bibitem{iac}
\BIBentryALTinterwordspacing
 [Online]. Available: \url{https://www.indyautonomouschallenge.com/}
\BIBentrySTDinterwordspacing

\bibitem{buehler20072005}
M.~Buehler, K.~Iagnemma, and S.~Singh, \emph{The 2005 DARPA grand challenge: the great robot race}.\hskip 1em plus 0.5em minus 0.4em\relax Springer, 2007, vol.~36.

\bibitem{buehler2009darpa}
------, \emph{The DARPA urban challenge: autonomous vehicles in city traffic}.\hskip 1em plus 0.5em minus 0.4em\relax springer, 2009, vol.~56.

\bibitem{architectural_principles}
J.~Lala and R.~Harper, ``Architectural principles for safety-critical real-time applications,'' \emph{Proceedings of the IEEE}, vol.~82, no.~1, pp. 25--40, 1994.

\bibitem{fault_tolerance_control}
F.~Y. Chemashkin and A.~A. Zhilenkov, ``Fault tolerance control in cyber-physical systems,'' in \emph{2019 IEEE Conference of Russian Young Researchers in Electrical and Electronic Engineering (EIConRus)}, 2019, pp. 1169--1171.

\bibitem{fault_detection_kalman}
B.~Pourbabaee, N.~Meskin, and K.~Khorasani, ``Sensor fault detection, isolation, and identification using multiple-model-based hybrid kalman filter for gas turbine engines,'' \emph{IEEE Transactions on Control Systems Technology}, vol.~24, 05 2015.

\bibitem{model_free_detection}
C.~Alippi, S.~Ntalampiras, and M.~Roveri, ``Model-free fault detection and isolation in large-scale cyber-physical systems,'' \emph{IEEE Transactions on Emerging Topics in Computational Intelligence}, vol.~1, no.~1, pp. 61--71, 2017.

\bibitem{online_error_detection}
K.~Ding, S.~Ding, A.~Morozov, T.~Fabarisov, and K.~Janschek, ``On-line error detection and mitigation for time-series data of cyber-physical systems using deep learning based methods,'' in \emph{2019 15th European Dependable Computing Conference (EDCC)}, 2019, pp. 7--14.

\bibitem{case_for_shm}
A.~N. Srivastava and J.~Schumann, ``The case for software health management,'' in \emph{2011 IEEE Fourth International Conference on Space Mission Challenges for Information Technology}, 2011, pp. 3--9.

\bibitem{apollo_open}
\BIBentryALTinterwordspacing
(2020) Open platform. [Online]. Available: \url{https://apollo.auto/developer.html}
\BIBentrySTDinterwordspacing

\bibitem{openpilot}
\BIBentryALTinterwordspacing
(2021) comma.ai. [Online]. Available: \url{https://comma.ai/}
\BIBentrySTDinterwordspacing

\bibitem{autoware_auto}
\BIBentryALTinterwordspacing
(2021) Autoware.auto. [Online]. Available: \url{https://autowarefoundation.gitlab.io/autoware.auto/AutowareAuto/}
\BIBentrySTDinterwordspacing

\bibitem{autoware_sys_monitor}
\BIBentryALTinterwordspacing
(2021) System monitor for autoware. [Online]. Available: \url{https://tier4.github.io/autoware.iv/tree/main/system/system\_monitor/}
\BIBentrySTDinterwordspacing

\bibitem{guardauto}
K.~Cheng, Y.~Zhou, B.~Chen, R.~Wang, Y.~Bai, and Y.~Liu, ``Guardauto: {A} decentralized runtime protection system for autonomous driving,'' \emph{CoRR}, vol. abs/2003.12359, 2020.

\bibitem{SVD_fault_detection}
Y.~Jiang and S.~Yin, ``Recursive total principle component regression based fault detection and its application to vehicular cyber-physical systems,'' \emph{IEEE Transactions on Industrial Informatics}, vol.~14, no.~4, pp. 1415--1423, 2018.

\bibitem{multisensor_fault_detection}
S.~Safavi, M.~A. Safavi, H.~Hamid, and S.~Fallah, ``Multi-sensor fault detection, identification, isolation, and health forecasting for autonomous vehicles,'' \emph{Sensors (Basel)}, vol.~21, no.~7, pp. 2547--2569, Apr 2021.

\bibitem{hybrid_approaches}
Y.~Fang, H.~Min, W.~Wang, Z.~Xu, and X.~Zhao, ``A fault detection and diagnosis system for autonomous vehicles based on hybrid approaches,'' \emph{IEEE Sensors Journal}, vol.~20, no.~16, pp. 9359--9371, 2020.

\bibitem{Safe_Motion_Planning}
\BIBentryALTinterwordspacing
A.~Liniger and L.~van Gool, ``Safe motion planning for autonomous driving using an adversarial road model,'' 2020. [Online]. Available: \url{https://arxiv.org/abs/2005.07691}
\BIBentrySTDinterwordspacing

\bibitem{IAC_TUM}
\BIBentryALTinterwordspacing
J.~Betz, T.~Betz, F.~Fent, M.~Geisslinger, A.~Heilmeier, L.~Hermansdorfer, T.~Herrmann, S.~Huch, P.~Karle, M.~Lienkamp, B.~Lohmann, F.~Nobis, L.~Ögretmen, M.~Rowold, F.~Sauerbeck, T.~Stahl, R.~Trauth, F.~Werner, and A.~Wischnewski, ``{TUM} autonomous motorsport: An autonomous racing software for the indy autonomous challenge,'' \emph{Journal of Field Robotics}, jan 2023. [Online]. Available: \url{https://doi.org/10.1002\%2Frob.22153}
\BIBentrySTDinterwordspacing

\bibitem{IAC_KAIST}
\BIBentryALTinterwordspacing
D.~Lee, C.~Jung, A.~Finazzi, H.~Seong, and D.~H. Shim, ``A resilient navigation and path planning system for high-speed autonomous race car,'' 2022. [Online]. Available: \url{https://arxiv.org/abs/2207.12232}
\BIBentrySTDinterwordspacing

\bibitem{IAC_EURO}
\BIBentryALTinterwordspacing
A.~Raji, A.~Liniger, A.~Giove, A.~Toschi, N.~Musiu, D.~Morra, M.~Verucchi, D.~Caporale, and M.~Bertogna, ``Motion planning and control for multi vehicle autonomous racing at high speeds,'' 2022. [Online]. Available: \url{https://arxiv.org/abs/2207.11136}
\BIBentrySTDinterwordspacing

\bibitem{IAC_MIT}
\BIBentryALTinterwordspacing
J.~Spisak, A.~Saba, N.~Suvarna, B.~Mao, C.~T. Zhang, C.~Chang, S.~Scherer, and D.~Ramanan, ``Robust modeling and controls for racing on the edge,'' 2022. [Online]. Available: \url{https://arxiv.org/abs/2205.10841}
\BIBentrySTDinterwordspacing

\bibitem{Stahl_RoboRace1}
\BIBentryALTinterwordspacing
T.~Stahl, M.~Eicher, J.~Betz, and F.~Diermeyer, ``Online verification concept for autonomous vehicles -- illustrative study for a trajectory planning module,'' 2020. [Online]. Available: \url{https://arxiv.org/abs/2005.07740}
\BIBentrySTDinterwordspacing

\bibitem{Stahl_RoboRace2}
T.~Stahl and F.~Diermeyer, ``Online verification enabling approval of driving functions—implementation for a planner of an autonomous race vehicle,'' \emph{IEEE Open Journal of Intelligent Transportation Systems}, vol.~2, pp. 97--110, 2021.

\bibitem{ROS_rescue}
P.~Kaveti and H.~Singh, ``{ROS} rescue : Fault tolerance system for robot operating system,'' \emph{CoRR}, vol. abs/1910.01078, 2019.

\bibitem{performance_ros2}
Y.~Maruyama, S.~Kato, and T.~Azumi, ``Exploring the performance of ros2,'' in \emph{Proceedings of the 13th International Conference on Embedded Software}, ser. EMSOFT '16.\hskip 1em plus 0.5em minus 0.4em\relax New York, NY, USA: Association for Computing Machinery, 2016.

\bibitem{robotic_anomaly_detection}
R.~Gupta, Z.~T. Kurtz, S.~A. Scherer, and J.~M. Smereka, ``Open problems in robotic anomaly detection,'' \emph{CoRR}, vol. abs/1809.03565, 2018.

\bibitem{8806893}
H.~Tabani, L.~Kosmidis, J.~Abella, F.~J. Cazorla, and G.~Bernat, ``Assessing the adherence of an industrial autonomous driving framework to iso 26262 software guidelines,'' in \emph{2019 56th ACM/IEEE Design Automation Conference (DAC)}, 2019, pp. 1--6.

\bibitem{10.1007/978-3-030-26601-1_2}
T.~Ishigooka, S.~Otsuka, K.~Serizawa, R.~Tsuchiya, and F.~Narisawa, ``Graceful degradation design process for autonomous driving system,'' in \emph{Computer Safety, Reliability, and Security}, A.~Romanovsky, E.~Troubitsyna, and F.~Bitsch, Eds.\hskip 1em plus 0.5em minus 0.4em\relax Cham: Springer International Publishing, 2019, pp. 19--34.

\bibitem{pure_pursuit}
M.~Park, S.~Lee, and W.~Han, ``Development of steering control system for autonomous vehicle using geometry-based path tracking algorithm,'' \emph{Etri Journal}, vol.~37, no.~3, pp. 617--625, 2015.

\bibitem{mpc}
M.~Nolte, M.~Rose, T.~Stolte, and M.~Maurer, ``Model predictive control based trajectory generation for autonomous vehicles — an architectural approach,'' in \emph{2017 IEEE Intelligent Vehicles Symposium (IV)}, 2017, pp. 798--805.

\bibitem{ekf}
R.~E. Kopp and R.~J. Orford, ``Linear regression applied to system identification for adaptive control systems,'' \emph{AIAA JOURNAL}, vol.~1, no.~10, pp. 2300--2306, Oct 1963.

\bibitem{plane_detection}
J.~Poppinga, N.~Vaskevicius, A.~Birk, and K.~Pathak, ``Fast plane detection and polygonalization in noisy 3d range images,'' in \emph{2008 IEEE/RSJ International Conference on Intelligent Robots and Systems}, 2008, pp. 3378--3383.

\bibitem{euclidean_clustering}
M.~Ester, H.-P. Kriegel, J.~Sander, and X.~Xu, ``A density-based algorithm for discovering clusters in large spatial databases with noise,'' in \emph{Proceedings of the Second International Conference on Knowledge Discovery and Data Mining}, ser. KDD'96.\hskip 1em plus 0.5em minus 0.4em\relax AAAI Press, 1996, p. 226–231.

\bibitem{fmeca_nasa}
R.~J. Duphily and A.~F.~S. Command, ``Space vehicle failure modes, effects, and criticality analysis (fmeca) guide,'' \emph{Space Missile Syst. Center, El Segundo, CA, USA, Aerosp. Rep. No. TOR-2009 (8591)-13}, 2009.

\bibitem{fmeca_car}
J.~Gong, Y.~Luo, Z.~Qiu, and X.~Wang, ``Determination of key components in automobile braking systems, based on abc classification and fmeca,'' \emph{Journal of Traffic and Transportation Engineering (English Edition)}, vol.~9, 06 2020.

\bibitem{fmeca_autonomous}
D.~Crestani, K.~Godary-Dejean, and L.~Lapierre, ``Enhancing fault tolerance of autonomous mobile robots,'' \emph{Robotics and Autonomous Systems}, vol.~68, 01 2015.

\bibitem{driscoll2003byzantine}
K.~Driscoll, B.~Hall, H.~Sivencrona, and P.~Zumsteg, ``Byzantine fault tolerance, from theory to reality,'' in \emph{Computer Safety, Reliability, and Security: 22nd International Conference, SAFECOMP 2003, Edinburgh, UK, September 23-26, 2003. Proceedings 22}.\hskip 1em plus 0.5em minus 0.4em\relax Springer, 2003, pp. 235--248.

\end{thebibliography}
